\documentclass[10pt,twocolumn,letterpaper]{article}

\usepackage{iccv}
\usepackage{times}
\usepackage{epsfig}
\usepackage{graphicx}
\usepackage{amsmath}
\usepackage{amssymb}
\usepackage{tabularx}
\usepackage{subcaption}
\usepackage{multicol} 
\usepackage{multirow} 
\usepackage{booktabs}
\usepackage{mathabx}
\usepackage[export]{adjustbox}
\usepackage{afterpage}
\usepackage{url}            
\def\UrlBreaks{\do\/\do-}
\DeclareRobustCommand{\quartiles}[3]{#2\hspace{1pt}\rlap{\textsubscript{#1}}{\textsuperscript{\raisebox{1pt}{#3}}}}
\usepackage{algorithm}
\usepackage[noend]{algpseudocode}

\DeclareMathOperator*{\argmin}{arg\,min}
\usepackage[export]{adjustbox}
\DeclareRobustCommand{\quartiles}[3]{#2\hspace{1pt}\rlap{\textsubscript{#1}}{\textsuperscript{\raisebox{1pt}{#3}}}}
\usepackage{algorithm}
\usepackage{mathabx}
\usepackage{url}            
\usepackage[noend]{algpseudocode}
\usepackage[pagebackref=true,breaklinks=true,colorlinks,bookmarks=false]{hyperref}
\usepackage{MnSymbol,wasysym}
\newcommand{\settitle}{\@maketitle}

\iccvfinalcopy 

\def\iccvPaperID{3864} 

\ificcvfinal\fi

\begin{document}

\title{PICCOLO: Point Cloud-Centric Omnidirectional Localization}

\author{Junho Kim, Changwoon Choi, Hojun Jang, and Young Min Kim\\
Dept. of Electrical and Computer Engineering, Seoul National University, Korea\\
{\tt\small 82magnolia@snu.ac.kr, changwoon.choi00@gmail.com, \{j12040208, youngmin.kim\}@snu.ac.kr}
}

\maketitle
\ificcvfinal\thispagestyle{empty}\fi

\begin{abstract}
We present PICCOLO, a simple and efficient algorithm for omnidirectional localization. 
Given a colored point cloud and a $360^\circ$ panorama image of a scene, our objective is to recover the camera pose at which the panorama image is taken.
Our pipeline works in an off-the-shelf manner with a single image given as a query and does not require any training of neural networks or collecting ground-truth poses of images.
Instead, we match each point cloud color to the holistic view of the panorama image with gradient-descent optimization to find the camera pose.
Our loss function, called sampling loss, is point cloud-centric, evaluated at the projected location of every point in the point cloud.
In contrast, conventional photometric loss is image-centric, comparing colors at each pixel location.
With a simple change in the compared entities, sampling loss effectively overcomes the severe visual distortion of omnidirectional images, and enjoys the global context of the $360^\circ$ view to handle challenging scenarios for visual localization.
PICCOLO outperforms existing omnidirectional localization algorithms in both accuracy and stability when evaluated in various environments.
Code is available at \url{https://github.com/82magnolia/panoramic-localization/}.

\end{abstract}

\section{Introduction}

\begin{figure}[t]
\centering\includegraphics[width=\linewidth]{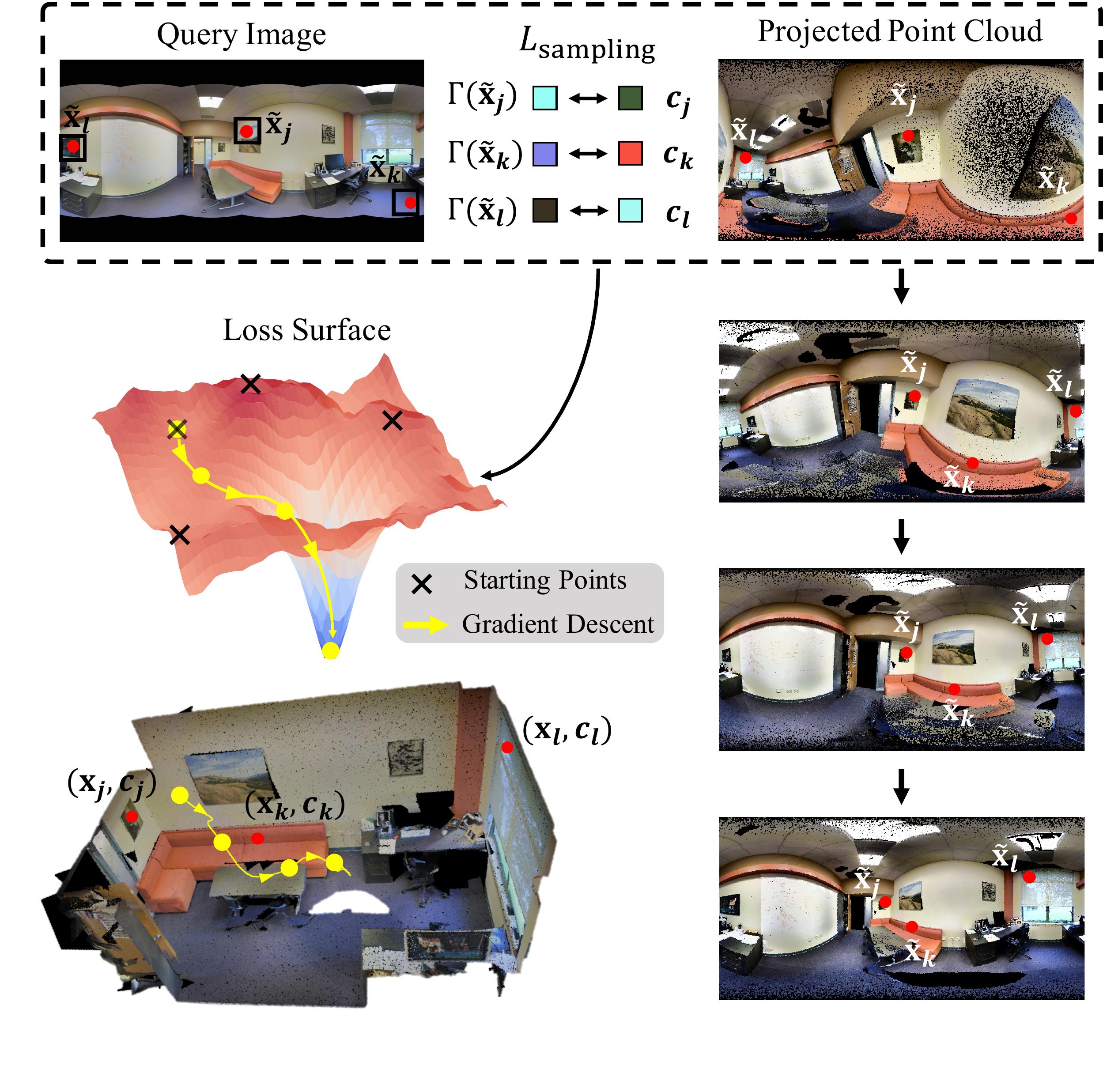}
\caption{Overview of our approach. PICCOLO minimizes a novel, point cloud-centric loss function called sampling loss. 
After the initialization phase trims off local minima, PICCOLO minimizes the sampling loss with gradient descent.
}
\label{impactful_demo}
\vspace{-0.5em}
\end{figure}

With the recent advancements in 3D sensing technology, 3D maps of the environment are often available for download~\cite{matterport_scan} or can be easily captured with commodity sensors~\cite{bundlefusion}.
The 3D map and the accurate location of the user within the map provide crucial information for AR/VR applications or other location-based services.
Visual localization is a cheap localization method as it only uses an image input and utilizes the 3D map without additional sensors such as WIFI, GPS, or gyroscopes.
However, visual localization is fragile to changes in illumination or local geometric variations resulting from object displacements~\cite{change_vis_1, change_vis_2}. 
Further, with the limited field of view, perspective cameras often fail to regress the camera pose when the observed image lacks visual features (e.g., a plain wall) or the scene exhibits symmetric or repetitive structure~\cite{lstm_vis_loc, inloc}.

Omnidirectional cameras, equipped with a $360^\circ$ field of view, provide a holistic view of the surrounding environment.
Hence these cameras are immune to small scene changes and ambiguous local features~\cite{benefits_of_omni}, which gives them the potential to dramatically improve the performance of visual localization algorithms.
However, the large field of view comes with a cost: significant visual distortion caused by the spherical projection equation.
This makes it difficult to directly apply conventional visual localization algorithms on omnidirectional cameras~\cite{omni_hard_why_1, omni_hard_why_2, omni_hard_why_3, omni_hard_why_4, spherenet}, as many visual localization algorithms~\cite{da4ad, inloc, lstm_vis_loc, score} do not account for distortion.
Furthermore, learning-based approaches are bound to the settings they are trained on, and cannot generalize to arbitrary scenes.

\begin{figure*}[h]
\makebox[\textwidth][c]{
    \includegraphics[width=0.24\linewidth]{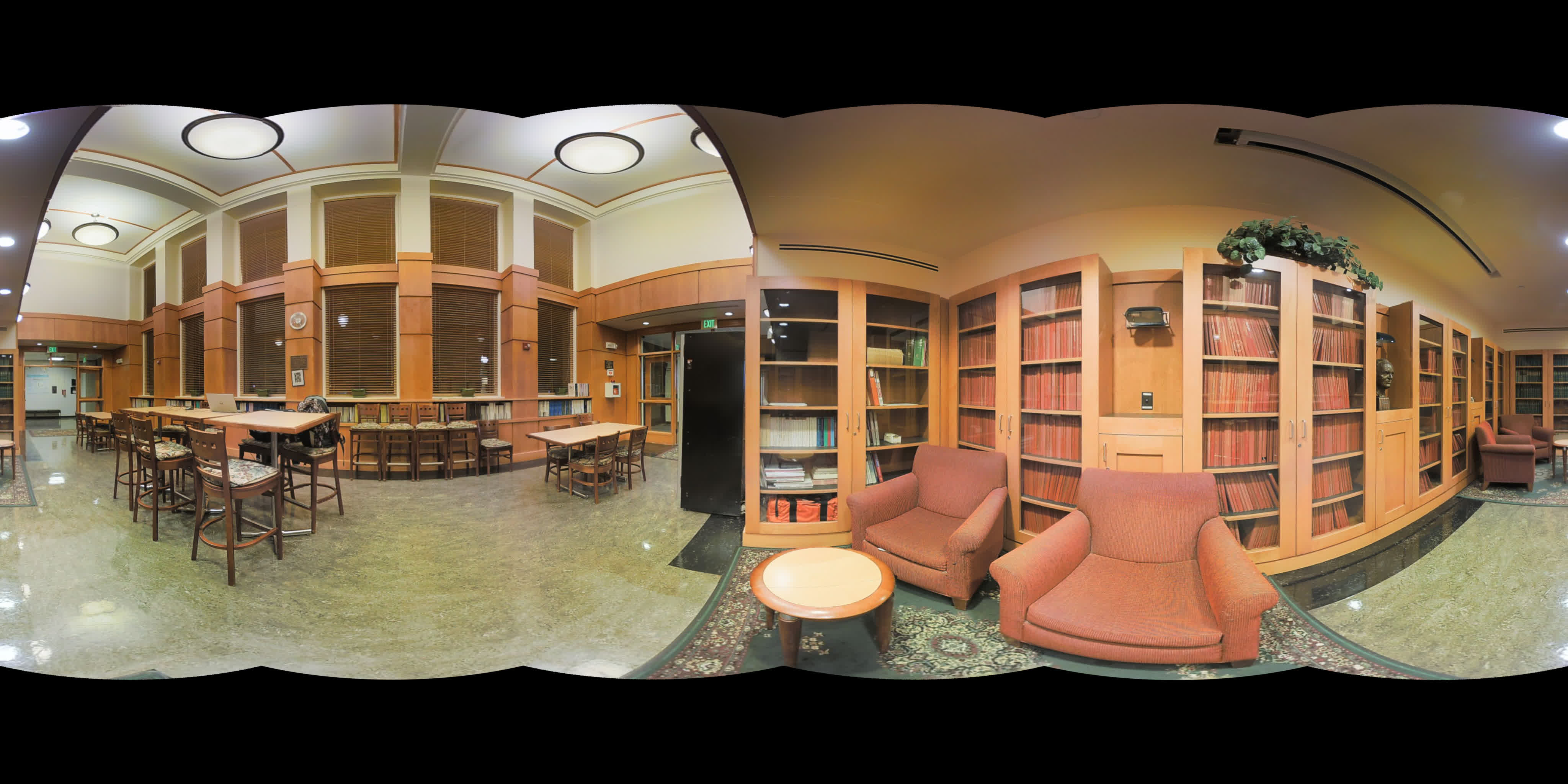}
    \includegraphics[width=0.24\linewidth]{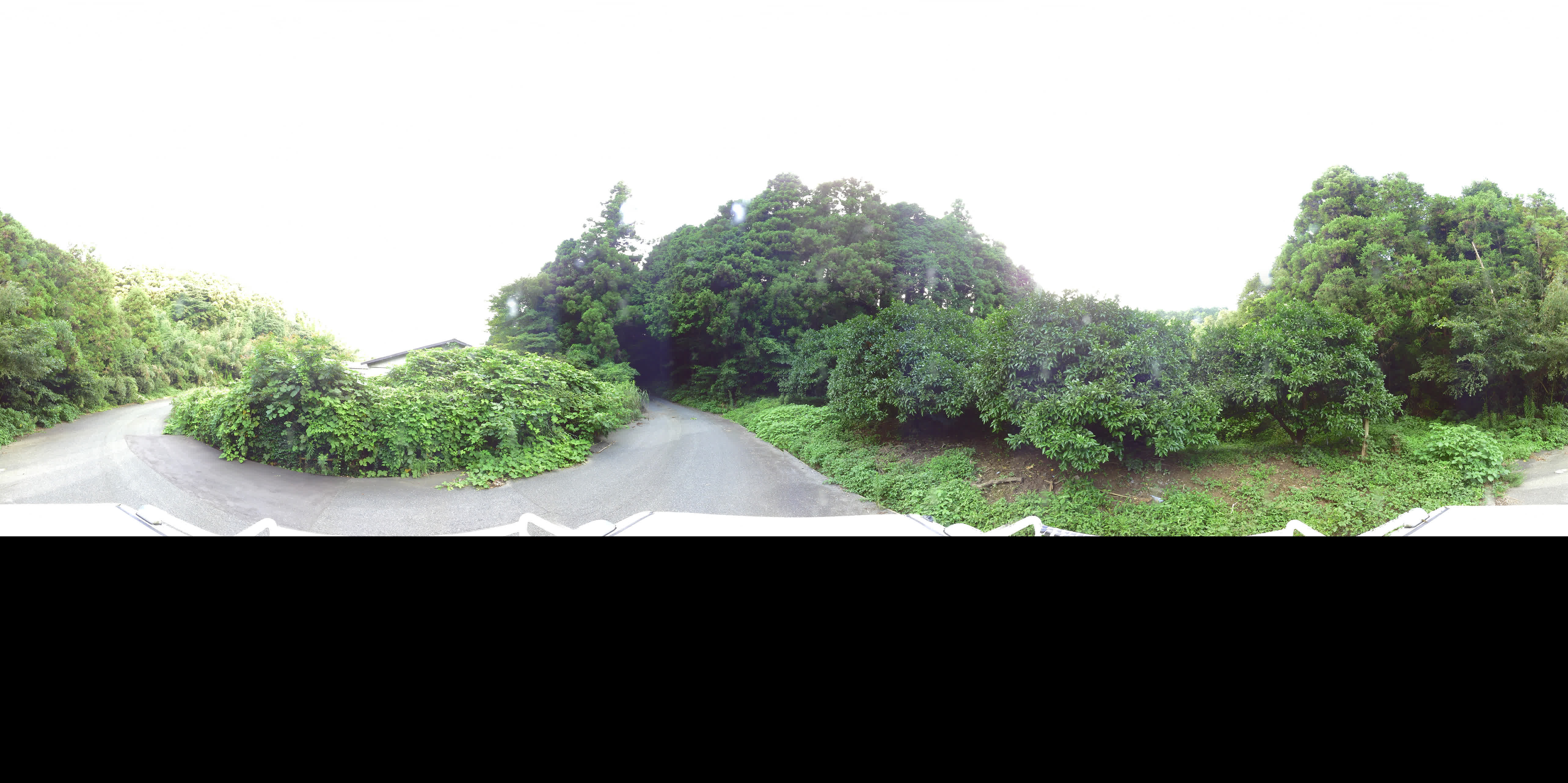}
    \includegraphics[width=0.24\linewidth]{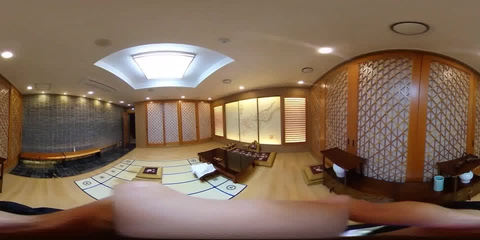}
    \includegraphics[width=0.24\linewidth]{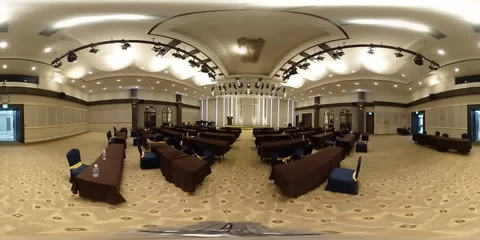}
}

\makebox[\textwidth][c]{
    \includegraphics[width=0.24\linewidth]{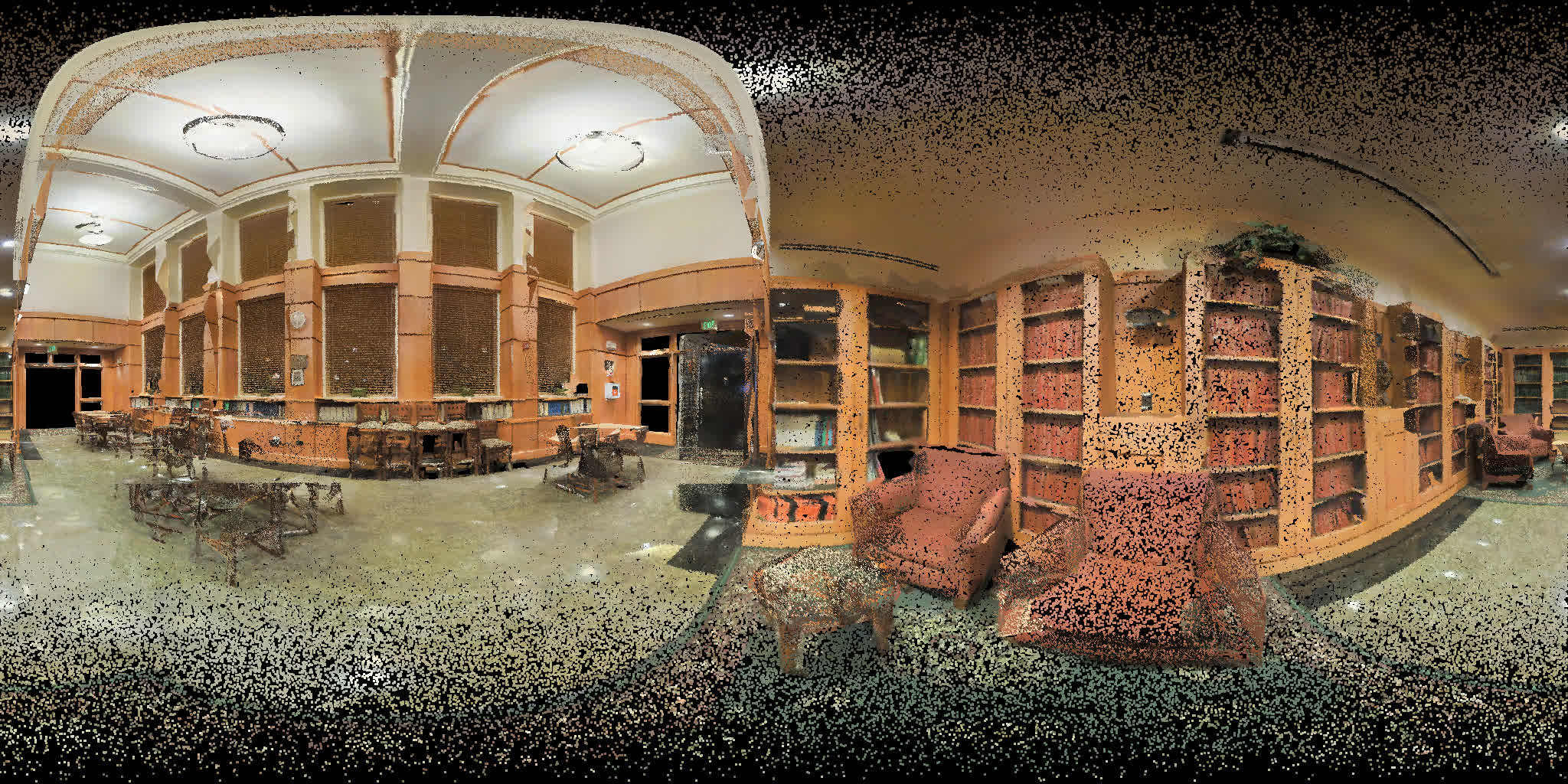}
    \includegraphics[width=0.24\linewidth]{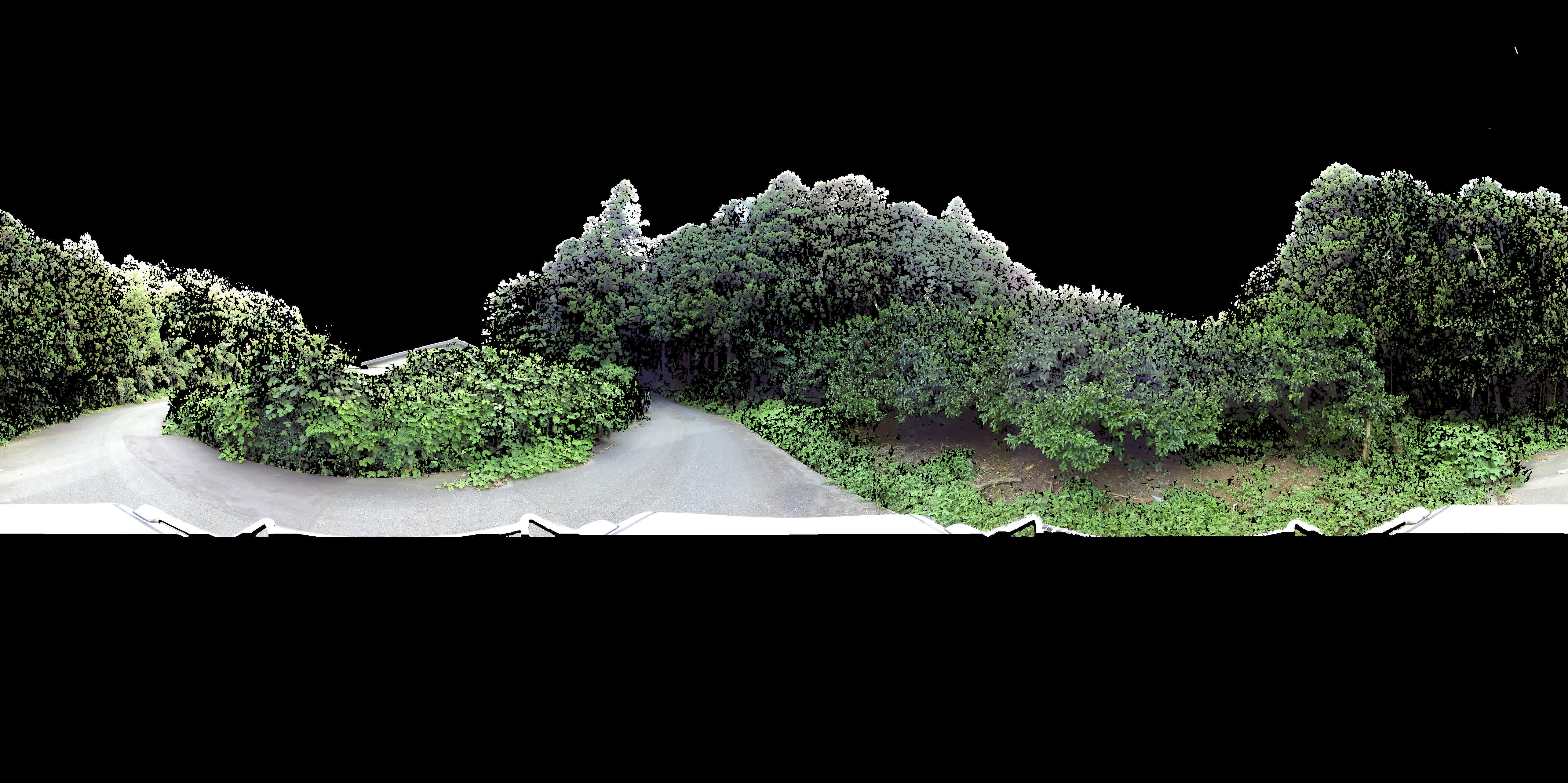}
    \includegraphics[width=0.24\linewidth]{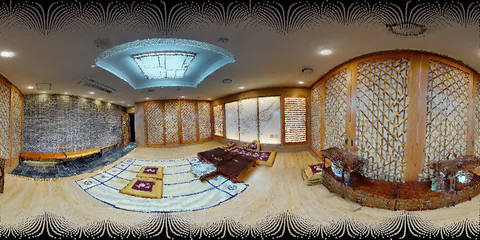}
    \includegraphics[width=0.24\linewidth]{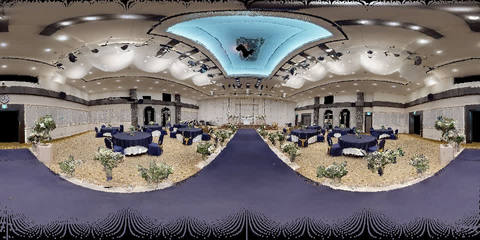}
}

\makebox[\textwidth][c]{
    \makebox[0.24\textwidth][c]{
        \textit{Stanford2D-3D-S}
    }
    \makebox[0.24\textwidth][c]{
        \textit{MPO}
    }
    \makebox[0.24\textwidth][c]{
        \textit{OmniScenes}
    }
    \makebox[0.24\textwidth][c]{
    \textit{OmniScenes (w. change)}
    }
}
\caption{Qualitative results of PICCOLO. We display the input query image (top), and the projected point cloud under the estimated camera pose (bottom).}
\label{teaser_img}
\end{figure*}

In this paper, we introduce PICCOLO, a simple yet effective omnidirectional localization algorithm.
PICCOLO optimizes over \textit{sampling loss}, which samples color values from the query image and compares them with the point cloud color.
We only utilize the color information from point clouds, as it is usually available from raw measurements.
With a simple formulation, PICCOLO can be adapted to any scene with 3D maps in an off-the-shelf manner.
Further, PICCOLO can work seamlessly with any other point-wise information, such as semantic segmentation labels shown in Figure~\ref{qualitative_results}.
Sampling loss is \textit{point cloud-centric}, as every point is taken into consideration.
In contrast, conventional photometric loss widely used in computer vision~\cite{flownet, lsd_slam} evaluates the color difference at every pixel location~\cite{lsd_slam, dtam}, thus is \textit{image-centric}.
Our point cloud-centric formulation leads to a significant performance boost in omnidirectional localization, where the image-centric approach suffers from distorted omnidirectional images unless the distortion is explicitly considered with additional processing~\cite{sphere_cnn,omni_hard_why_4}.

The gradient of our proposed sampling loss can be efficiently obtained with differentiable sampling~\cite{spatial_transformer}.
While differentiable sampling is widely used to minimize discrepancies in the projected space, it is usually part of a learned module~\cite{flownet, flownet2}.
Instead, we utilize the operation in a stand-alone fashion, making our framework cheap to compute.
We further accelerate the loss computation by ignoring the non-differentiable, costly components of projection, such as occlusion handling.
These design choices make sampling loss very fast: it only takes 3.5 ms for $10^6$ points on a commodity GPU.
With the rich information of the global context in point cloud color, our efficient formulation is empirically robust against visual distortions and more importantly, local scene changes.
The algorithm quickly converges to the global minimum of the proposed loss function as shown in Figure~\ref{impactful_demo}.

Equipped with a light-weight search for decent starting points, PICCOLO achieves stable localization in various datasets.
The algorithm is extensively evaluated on indoor/outdoor scenes and scenes with dynamic camera motion, scene changes, and arbitrary point cloud rotation.
Several qualitative results of our algorithm are shown in Figure~\ref{teaser_img} and~\ref{qualitative_results}.
In addition, we introduce a new dataset called OmniScenes to highlight the practicality of PICCOLO.
OmniScenes contains diverse recordings with significant scene changes and motion blur, making it the first dataset targeted for omnidirectional localization where visual localization algorithms frequently malfunction.
PICCOLO consistently exhibits performance superior to the previous approaches~\cite{gosma, sphere_cnn} in all of the tested datasets under a fixed hyperparameter configuration, indicating the practical effectiveness of our algorithm.

\section{Related Work}
\label{related_work}

Before we introduce PICCOLO in detail, we clarify our problem setup and how it differs from previous visual localization algorithms~\cite{active_search_eccv, inloc, energy_landscape, dsac}.
Then we will further describe recent algorithms proposed for omnidirectional localization.
 
\paragraph{Learning-based Algorithms} 
A large body of recent visual localization literature trains an algorithm on the database of RGB (and possibly depth) images annotated  with ground truth poses~\cite{energy_landscape, dsac, hierarchical_scene, score, inloc, posenet, lstm_vis_loc, learned_loc_1, learned_loc_2, learned_loc_3, learned_loc_4}.
While such training facilitates highly accurate camera pose estimation~\cite{energy_landscape, dsac, inloc, active_search}, it limits the applicability of these algorithms.
To estimate camera pose in new, unseen environments, these algorithms typically require additional pose-labelled samples.

In order to develop an algorithm that could be readily used in an off-the-shelf manner, we make a slight detour from these previous setups: the camera pose must be found \emph{solely} using the point cloud and query image information.
One may opt to train these learning-based models~\cite{energy_landscape, dsac, hierarchical_scene} with synthesized views from the point cloud as in Zhang \etal~\cite{sphere_cnn}.
However, it is costly to obtain such rendered views, and one must devise a way to reduce the domain gap between synthesized images and real query images, which is a non-trivial task.

\paragraph{Feature-based algorithms} 
Another line of work utilizes visual features for localization~\cite{active_search_eccv, active_search, snavely, synthetic_view_with_matching, loc_feat_1, loc_feat_2}.
Feature-based localization algorithms require each 3D point to be associated with a visual feature, typically SIFT~\cite{sift}, necessitating a structure-from-motion (SfM) point cloud.
Provided an efficient search scheme~\cite{active_search, active_search_eccv, snavely}, it is relatively straightforward to establish 2D-3D correspondences by matching features extracted from the query image with those in the SfM model.

Our input point cloud is not limited to a structure-from-motion (SfM) point cloud.
Due to the developments in RGB-D sensors and Lidar scanners, 3D point clouds of a scene could be obtained in a wide variety of ways other than SfM.
These point clouds do not contain associated visual features for feature-based localization.
Our setup also does not provide any explicit 2D-3D correspondences, thus disabling the direct usage of PnP algorithms~\cite{pnp_1, pnp_2}.
Further, many point clouds and query images used in our experiments contain repetitive structures or regions that lack features as shown in Figure~\ref{qualitative_results}.
This hinders the usage of sparse local features such as SIFT~\cite{sift} in our setup, where we report additional difficulties for using SIFT in the supplementary material.
To accommodate these challenges, PICCOLO incorporates information from dense RGB measurements, which are easy to obtain in practice, and are robust against local ambiguities.

\paragraph{Omnidirectional Localization}
Visual localization on omnidirectional images requires an algorithm specifically designed to account for the unique visual distortion~\cite{omni_hard_why_1, omni_hard_why_2, omni_hard_why_3, omni_hard_why_4}.
A number of techniques have been proposed in recent years that tackle visual localization with omnidirectional cameras.
These techniques could be divided into two groups, namely algorithms that utilize global optimization techniques and others that leverage deep learning.
Campbell \etal \cite{gopac, gosma} proposed a family of global optimization-based algorithms for camera pose estimation, GOSMA \cite{gosma} and GOPAC \cite{gopac}, that could be readily applied for omnidirectional localization in diverse indoor and outdoor environments.
While these algorithms have solid optimality guarantees and competitive performance, semantic labels should be fed to these algorithms as additional inputs for reasonable accuracy.
On the other hand, deep learning-based omnidirectional localization algorithms such as Zhang \etal\cite{sphere_cnn}, train neural networks that learn rotationally equivariant features to effectively process omnidirectional images.
Although such features enable omnidirectional localization under arbitrary camera rotations, these algorithms cannot generalize to unobserved scenes as they require training on pose-annotated images. 
We compare the localization performance of PICCOLO with optimization-based localization algorithms GOSMA \cite{gosma}, GOPAC \cite{gopac}, and deep learning-based localization algorithms from Zhang \etal \cite{posenet, sphere_cnn, spherenet}.
\section{Method}
\label{method}

PICCOLO is a point cloud-centric omnidirectional localization algorithm, which finds the optimal $SE(3)$ camera pose with respect to the colored point cloud at which the $360^{\circ}$ panorama image is taken.
PICCOLO solely relies on the point cloud data and the input query image.
It does not require a separate training process or explicit 2D-3D correspondences, and therefore could be used in an off-the-shelf manner.
We first introduce the formulation of sampling loss, which is the objective function that PICCOLO aims to minimize.
Then we will describe our light-weight initialization scheme.

\begin{figure*}[t]
\centering
\includegraphics[width=0.85\linewidth]{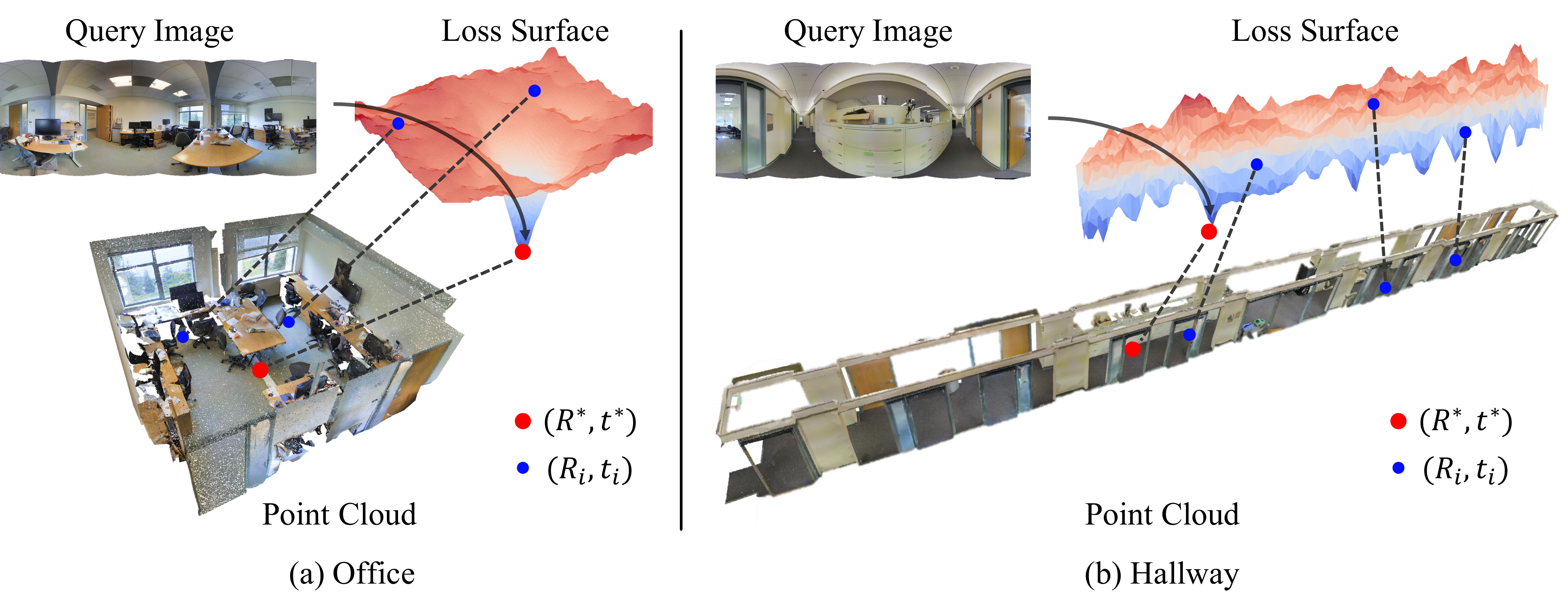}
\caption{Visualization of loss surfaces obtained from scenes in the Stanford2D-3D-S dataset~\cite{stanford2d3d}. 
The loss surfaces show the minimum loss values of the given ($x,y$) position in the 3D space.
The red dots indicate the ground truth camera positions, and the blue dots link the values on the loss surface and the corresponding camera positions within the input point cloud space.
Loss surfaces of small scenes are typically smooth with clear global minimum (left), but those of large scenes contain numerous local minima (right).}
\vspace{-0.5em}
\label{loss_surfaces}
\vspace{-0.5em}
\end{figure*}

\paragraph{Sampling Loss}
Given a point cloud $P=\{X, C\}$ and a \emph{single} query image $I \in \mathbb{R}^{H\times W \times 3}$, where $X, C \in \mathbb{R}^{N\times 3}$ are the point cloud coordinates and color values, the objective is to find the optimal rotation $R^* \in SO(3)$ and translation $t^* \in \mathbb{R}^{3}$ at which the $360^{\circ}$ panorama image $I$ is taken.
Denote $\Pi(\cdot):\mathbb{R}^{3}\rightarrow \mathbb{R}^{2}$ as the projection function that maps a point $\mathbf{x}=(x_1, x_2, x_3)$ in 3D to a point $\tilde{\mathbf{x}} \in [0,H)\times[0,W)$ in the $360^{\circ}$ panorama image's coordinate frame.
This could be explicitly written as follows,
\begin{equation}
\Pi(\mathbf{x}) = 
\bigg(\frac{H}{\pi}\mathrm{atan}\bigg(\frac{x_3}{\sqrt{x_1^2+x_2^2}}\bigg),
\frac{W}{2\pi}\mathrm{atan}\bigg(\frac{x_2}{x_1}\bigg)\bigg).
\end{equation}
Furthermore, let $\Gamma(\cdot;I)$ indicate the sampling function that maps 2D coordinates $\tilde{\mathbf{x}} \in [0,W)\times[0,H)$ to pixel values $\mathbf{c}\in \mathbb{R}^3$ sampled from the query image $I$ under a designated sampling kernel.
Suppose $\Gamma(\cdot;I),\Pi(\cdot)$ could be `vectorized', i.e., if the input $\tilde{X}$ consists of $N$ points in $\mathbb{R}^2$, $\Gamma(\tilde{X};I) \in \mathbb{R}^{N\times 3}$ are the sampled image values at 2D coordinates $\tilde{X}$, and vice versa for $\Pi(\cdot)$.

Under this setup, $\Pi(X) \in \mathbb{R}^{N\times 2}$ could be regarded as tentative \emph{sampling locations}, and $\Gamma(\Pi(X);I) \in \mathbb{R}^{N\times 3}$ as the \emph{sampled image values}.
If the point cloud $P$ is perfectly aligned with the omnidirectional camera's coordinate frame, one could expect the sampled image values $\Gamma(\Pi(X);I)$ to be very close to the point cloud color values $C$.
Sampling loss is derived from this observation, where the objective is to minimize the discrepancy between $\Gamma(\Pi(X);I)$ and $C$.
Given a candidate camera pose $R, t$, this could be formulated as follows, 
\begin{equation}
    \label{samp_loss}
    L_\mathrm{sampling}(R, t) = \|\Gamma(\Pi(R(X-t));I) - C\|_2.
\end{equation}
Note that $R(X - t)$ is the transformed point cloud under $R, t$.
Gradients with respect to $R, t$ could be obtained by differentiating through the sampling function $\Gamma(\cdot;I)$ using the technique from Jaderberg \etal~\cite{spatial_transformer}.
Once the gradients are known, any off-the-shelf gradient based optimization algorithm such as stochastic gradient descent \cite{sgd} or Adam \cite{adam} could be applied to minimize Equation~\ref{samp_loss}, as shown in Figure~\ref{impactful_demo}. 

Unlike photometric loss which stems from an \emph{image-centric} viewpoint, sampling loss aims at assigning an adequate sampled color value to each point in the point cloud, thus providing a \emph{point cloud-centric} viewpoint.
Specifically, photometric loss also compares the colors of the point cloud with the query image, but in the image space, namely,
\begin{equation}
\label{second_photo}
    L_\mathrm{photometric}(R, t) = \|\Psi(\{R(X-t), C\}) - I\|_2,
\end{equation}
where $\Psi(\cdot):\{\mathbb{R}^{N\times3},\mathbb{R}^{N\times3}\} \rightarrow \mathbb{R}^{H\times W\times 3}$ is a rendering function that receives the point cloud to produce a synthesized image.
The rendering function is necessary to apply photometric loss in our setup, as only a single image is given, unlike existing applications~\cite{lsd_slam, dtam} where multiple images are provided.
As photometric loss is evaluated in the image space, it suffers from the visual distortion of omnidirectional cameras.
To illustrate, in Figure~\ref{teaser_img}, one can observe that points near the pole (ceilings, floors) are `stretched', while they correspond to small areas in reality.
Since photometric loss makes direct image comparisons, it is severely affected by such artifacts and requires additional processing to account for the distortion~\cite{sphere_cnn, omni_hard_why_4}.

Sampling loss has numerous advantages over photometric loss.
First, as seen from Equation~\ref{samp_loss}, it fairly incorporates all points in the point cloud agnostic of whether it is closer to the pole, thus making it more suitable for 6-DoF pose estimation of omnidirectional cameras.
Second, sampling loss is cheap to compute, while still allowing for easy gradient computation~\cite{spatial_transformer}.
Each sampling operation consists of simple image indexing, and we ignore the non-differentiable, costly components of projection, such as occlusion handling.
The core part of PICCOLO consists of simple gradient descent on the sampling loss, which is very fast: $3 \times 10^8$ points can be processed per second.
Nonetheless, sampling loss can effectively handle the holistic view of the $360^\circ$ image and is robust to visual distortion or minor scene changes.

\label{sec:init}

\paragraph{Initialization Algorithm} While sampling loss has various amenable properties, it is non-convex as visualized in Figure~\ref{loss_surfaces}.
Depending on the initial position, optimization using gradient descent can stop at a local minimum, which can be a serious issue for large spaces.
To this end, we introduce a lightweight initialization algorithm, which outputs feasible starting points that are likely to yield global convergence.

We uniformly sample the space of possible camera positions, and filter them through a two-step selection process as presented in Algorithm~\ref{alg:PICCOLO}.
During the first step, we compute sampling loss values across $N_t \times N_r$ candidate camera poses and obtain the top $K_1$ smallest starting points (line 2).
Specifically, $N_t$ translations are chosen from the uniform grid on the point cloud bounding box, for which $N_r$ rotations, uniformly sampled from $SO(3)$, are selected.
Since sampling loss is very efficient, we can quickly compute the loss for all of the starting points.

Among the $K_1$ starting points, the second filtering process further selects $K_2$ $(K_2 \leq K_1)$ of them using color histogram intersections (line 3). 
Top $K_2$ candidate poses with the highest color distribution overlap with the query image are chosen.
Finally, the resulting $K_2$ starting points are individually optimized for a fixed number of iterations with respect to the sampling loss in $SE(3)$ (line 7).
At termination, the optimized camera pose  with the smallest sampling loss value is chosen (line 9). 

\begin{algorithm}
\caption{Overview of PICCOLO}
\textbf{Inputs:} Point cloud P = $\{X, C\}$, query image $I$ \\
\textbf{Output:} Camera pose $\hat{R}, \hat{t}$. 
    \begin{algorithmic}[1]
    \State $T \gets [(R_i, t_i) | i \in [1\mathrel{{.}\,{.}}\nobreak N_tN_r]]$ 
    \Comment{Starting points}
    \State $T \gets \texttt{\small getTopK}(\texttt{\small lossValue}(T,P,I), K_1)$
    \State $T \gets \texttt{\small getTopK}(\texttt{\small histIntersect}(T,P, I), K_2)$
    \State $V \gets [\:]$
    \ForAll {$(R_i, t_i) \in T$}
        \For {$\mathrm{iter} \in [1\mathrel{{.}\,{.}}\nobreak N_\mathrm{iter}]$}
            \State $(R_i, t_i) \gets (R_i, t_i) - \alpha\nabla L_\mathrm{sampling}(R_i, t_i)$
        \EndFor
        \State $V.\texttt{\small append}(L_\mathrm{sampling}(R_i, t_i))$
    \EndFor
    \State $(\hat{R}, \hat{t}) \gets \argmin_{R, t} V$
    \end{algorithmic}
\label{alg:PICCOLO}
\end{algorithm}

\section{Experimental Results}
\label{exp}

\subsection{Performance Analysis}
\label{perf}
\paragraph{Implementation Details}
PICCOLO is mainly implemented using PyTorch~\cite{pytorch}, and is accelerated with a single RTX 2080 GPU.
Once the starting point is selected as described in Section~\ref{sec:init}, we find the camera pose using Adam~\cite{adam} with step size $\alpha=0.1$ in all experiments.
PICCOLO is straightforward to implement, with the core part of the algorithm taking less than 10 lines of PyTorch code.
For results in which accuracy is reported, a prediction is considered correct if the  translation error is below $0.1$~m and the rotation error is below $5.0^\circ$.
All translation and rotation errors reported are median values, following the convention of \cite{gosma, gopac}.
The full hyperparameter setup and additional qualitative results are available in the supplementary material.

\begin{table}[]

\centering
\newcolumntype{C}{>{\centering\arraybackslash}X}
\renewcommand*{\arraystretch}{0.95}
\setlength{\tabcolsep}{2pt}
{\small
\begin{tabularx}{\columnwidth}{@{} l C C | C C @{}}
\toprule
Method & Information & Learning & $t$-error (m) & $R$-error ($^\circ$) \\ \midrule
PoseNet~\cite{posenet} & RGB & $\bigcirc$ & 2.41 & 28 \\
SphereNet~\cite{spherenet} & RGB & $\bigcirc$ & 2.29 & 26.7 \\
Zhang~\etal~\cite{sphere_cnn} & RGB & $\bigcirc$ & 1.64 & 9.15 \\
PICCOLO & RGB & $\bigtimes$ & \textbf{0.03} & \textbf{0.66} \\ \midrule
GOSMA~\cite{gosma} & Semantic & $\bigtimes$ & 1.27 & 51.44 \\
PICCOLO & Semantic & $\bigtimes$ & \textbf{0.01} & \textbf{0.28} \\ \bottomrule
\end{tabularx}
}
\caption{
	Quantitative results of omnidirectional localization evaluated on all areas of the Stanford2D-3D-S dataset~\cite{stanford2d3d}.
}
\label{table:stanford_all}
\end{table}

\begin{table}[t]
    \centering
    \newcolumntype{C}{>{\centering\arraybackslash}X}
    \renewcommand*{\arraystretch}{1.1}
    \setlength{\tabcolsep}{2pt}
    {\small

    	\begin{tabularx}{\columnwidth}{@{}l C C C C C@{}}
    	\toprule
    	 & PICCOLO & GOSMA & GOSMA\textsuperscript{-$\Lambda$} & GOPAC\\
    	\midrule
    	$t$-error (m) & \quartiles{0.00}{\textbf{0.01}}{0.07} & \quartiles{0.05}{0.08}{0.15} & \quartiles{0.09}{0.14}{0.23} & \quartiles{0.10}{0.15}{0.27}\\
    	$R$-error ($^{\circ}$) & \quartiles{0.11}{\textbf{0.21}}{0.56} & \quartiles{0.91}{1.13}{2.18} & \quartiles{1.25}{2.38}{4.61} & \quartiles{2.47}{3.78}{5.10}\\
    	\bottomrule
        \end{tabularx}
    }
    \caption{Localization results of PICCOLO, GOSMA, GOSMA without class labels (GOSMA\textsuperscript{-$\Lambda$}), and GOPAC for a subset of Area 3 from Stanford2D-3D-S~\cite{stanford2d3d}. \quartiles{$Q_1$}{$Q_2$}{$Q_3$} are quartile values of each metric. Results other than PICCOLO are excerpted from \cite{gosma}.}
    \label{table:stanford_area_3}
\end{table}

\paragraph{Stanford2D-3D-S}
We assess the localization performance of PICCOLO against existing methods using the Stanford2D-3D-S dataset~\cite{stanford2d3d}, as shown in Table~\ref{table:stanford_all} and~\ref{table:stanford_area_3}.
It is an indoor dataset composed of 1413 panoramic images subdivided into six different areas, and many scenes exhibit repetitive structure and lack visual features, as in Figure~\ref{qualitative_results}.
All areas are used for comparison except for GOPAC~\cite{gopac}, where we use a subset from Area 3 consisting of small rooms, as the algorithm's long runtime hinders large-scale evaluation.

PICCOLO outperforms all existing baselines by a large margin, showing an order-of-magnitude performance gain from its competitors.
GOSMA~\cite{gosma} and GOPAC~\cite{gopac} are optimization-based methods that do not utilize color measurements.
Instead, they require semantic labels for decent performance.
For fair comparisons with these algorithms, we make PICCOLO observe color-coded  semantic labels as input, as shown in Figure~\ref{qualitative_results}, and report the numbers in Table~\ref{table:stanford_all} (PICCOLO Semantic) and~\ref{table:stanford_area_3}.
Semantic labels lack visual features, thus finding camera pose in this setup is closer to solving a blind-PnP problem~\cite{softposit}.
However, PICCOLO operates seamlessly and outperforms GOSMA and GOPAC without the aid of rich visual information such as RGB inputs, consistently succeeding around 1~cm error.
Although GOSMA and GOPAC are powerful algorithms that guarantee global optimality, they often fail in large scenes such as hallways, where the qualitative results are shown in the supplementary material.

PICCOLO also shows superior performance against deep learning methods~\cite{sphere_cnn, spherenet, posenet}.
Nevertheless, it should be noted that there is a subtle distinction in the search spaces of these methods.
The translation domain for deep learning-based methods is the entire Stanford2D-3D-S dataset, while it is confined to a particular area for PICCOLO, similar to GOSMA~\cite{gosma}.
However, deep learning-based methods are given very strong prior information to cope with the large search space; they are trained on synthetic pose-annotated images, which are generated within 30 cm proximity of the test images.
This means the training images are very close to the ground truth.
Nevertheless, it must be acknowledged that deep learning methods are capable of regressing the pose at wider scales, about 5 times the maximum search scale ($1000~\mbox{m}^2$) attainable with PICCOLO (Table~\ref{table:fukuoka}).

\paragraph{MPO} 
Multi-Modal Panoramic 3D Outdoor (MPO)~\cite{fukuoka} dataset is an outdoor dataset which spans a large area ($1000\mbox{m}^2$) with many scenes containing repetition or lacking visual features.
As shown in Table~\ref{table:fukuoka}, PICCOLO performs competently with the same hyperparameter setting as Stanford2D-3D-S~\cite{stanford2d3d}, despite the large area of the dataset.
This validates our claim that PICCOLO could readily function as an off-the-shelf omnidirectional localization algorithm for both indoor/outdoor environments.

\begin{table}[t]
    \centering
    \newcolumntype{C}{>{\centering\arraybackslash}X}
    \renewcommand*{\arraystretch}{0.95}
    {\small
    \begin{tabularx}{\columnwidth}{lc|ccc}
        \toprule
        & Area ($m^2$) & $t$-error (m) & $R$-error ($^{\circ}$) & Acc.\\
        \midrule
        Coast & 458.0&0.79 & 2.18 & 0.40 \\
        Forest & 361.2&0.02 & 0.92 & 0.67 \\
        ParkingIn & 92.9&2.77 & 96.50 & 0.13 \\
        ParkingOut & 1381.2&1.74 & 9.77 & 0.07 \\
        Residential & 412.8&0.83 & 2.53 & 0.46 \\
        Urban& 1156.4&0.03 & 0.85 & 0.85 \\
        \midrule
        All & 646.3 & 0.80 & 2.10 & 0.45 \\
        \bottomrule
    \end{tabularx}
    }
    
    \caption{Localization error and accuracy of PICCOLO on Multi-Modal Panoramic 3D Outdoor (MPO) dataset~\cite{fukuoka}.}
    \label{table:fukuoka}
\end{table}
\begin{table}[t]
    \centering
    \newcolumntype{C}{>{\centering\arraybackslash}X}
    \renewcommand*{\arraystretch}{0.95}
    {\small
    \begin{tabularx}{\columnwidth}{lc|ccc}
        \toprule
        Scenario & Scene Change & $t$-error (m) & $R$-error ($^{\circ}$) & Acc.\\
        \midrule
        Handheld & $\bigtimes$ & 0.02 & 0.25 & 0.71\\
        Robot & $\bigtimes$ & 0.02 & 0.18 & 0.77\\
        \midrule
        Handheld & $\bigcirc$ & 0.77 & 15.39 & 0.43\\
        Robot & $\bigcirc$ & 0.05 & 0.59 & 0.55\\
        \bottomrule
    \end{tabularx}
    }
    
    \caption{Localization error and accuracy of PICCOLO on the OmniScenes dataset.}
    \label{table:omniscenes}
\end{table}

\paragraph{Practicality Assessment with OmniScenes}
Omnidirectional localization is expected to provide stable visual localization under scene changes or dynamic motion, and therefore promises practical applications in VR/AR or robotics.
We introduce a new dataset called OmniScenes collected to evaluate the performance on scenes with the aforementioned challenges.
We collect dense 3D scans of eight areas including wedding halls and hotel rooms using the Matterport3D Scanner~\cite{matterport_scan}.
Corresponding $360^\circ$ panoramic images are acquired with the Ricoh Theta $360^\circ$ camera~\cite{Ricoh} under two scenarios, handheld and mobile robot mounted.
Handheld scenarios are typically more challenging as unconstrained motion could take place and the capturer partially occludes scene details.
The images are taken at different times of day and include significant changes in furniture configurations and motion blurs.
Further details about the dataset are deferred to the supplementary material.

The evaluation results on the OmniScenes dataset are shown in Table~\ref{table:omniscenes}.
Unlike previous experiments, we assume that the gravity direction is known, as this is often available in practice.
PICCOLO exhibits competent error rates when there are no scene changes, agnostic of whether the input $360^\circ$ panorama is recorded in a handheld or robot-mounted manner.
As shown in Figure~\ref{qualitative_results}, PICCOLO can estimate camera pose even under severe handheld motion, thanks to the full incorporation of points from sampling loss.

Even though there is no functionality in PICCOLO that accounts for scene changes, there is a considerable amount of success cases given the accuracy in Table~\ref{table:omniscenes} and qualitative results shown in Figure~\ref{qualitative_results}.
As long as the global context provides enough amount of evidence from color samples, omnidirectional localization can succeed.
Nonetheless, there is a clear performance gap, and enhancing the robustness of PICCOLO against various scene changes is left as future work.

\subsection{Ablation Study}

In this section we ablate various components of PICCOLO. 
Experiments are conducted using all areas of the Stanford2D-3D-S dataset~\cite{stanford2d3d}, unless specified otherwise.

\paragraph{Sampling Loss}
We compare PICCOLO with a variant that uses photometric loss from Equation~\ref{second_photo} in place of the sampling loss, to ablate the effect of sampling loss in our algorithm.
The rendering function is implemented as a simple projection of the 3D point cloud, similar to projections shown in Figure~\ref{qualitative_results}.
We use the warping function to obtain gradients with respect to $R, t$, as in previous works~\cite{lsd_slam, dtam, flownet}.
All other hyperparameter setups and the initialization algorithm are the same as PICCOLO.

The design choice of using sampling loss shows a large performance gain over photometric loss, as shown in Table~\ref{table:photometric}.
As sampling loss fairly incorporates all points in point cloud, it is free from visual distortion and thus more suitable than photometric loss for 6-DoF omnidirectional localization.

\begin{table}[t]
    \centering
    \newcolumntype{C}{>{\centering\arraybackslash}X}
    \renewcommand*{\arraystretch}{0.95}
    {\small
    \begin{tabularx}{\columnwidth}{@{}lc|cc@{}}
        \toprule
        Loss Function & Information & $t$-error (m) & $R$-error ($^{\circ}$)\\
        \midrule
        Sampling & Original & \textbf{0.03} & \textbf{0.66} \\
        Photometric & Original & 1.41 & 42.29 \\
        \midrule
        Sampling & Gravity Direction & \textbf{0.01} & \textbf{0.34}\\
        Photometric & Gravity Direction & 0.93 & 33.41\\
        \midrule
        Sampling & Flipped & \textbf{0.03} & \textbf{0.69}\\
        Photometric & Flipped & 1.42 & 42.79\\
        \midrule
        Sampling & Rand. Rot. & \textbf{0.23} & \textbf{2.21}\\
        Photometric & Rand. Rot. & 1.48 & 43.33\\
        \bottomrule
    \end{tabularx}
    }
    
    \caption{Ablation study on sampling loss and gravity direction. `Flipped' denotes flipped query images and `Rand. Rot.' denotes randomly rotated point cloud inputs.}
    \label{table:photometric}
\end{table}
\begin{figure}[t]
\centering
    \begin{subfigure}[t]{1.0\linewidth}
    \includegraphics[width=\linewidth]{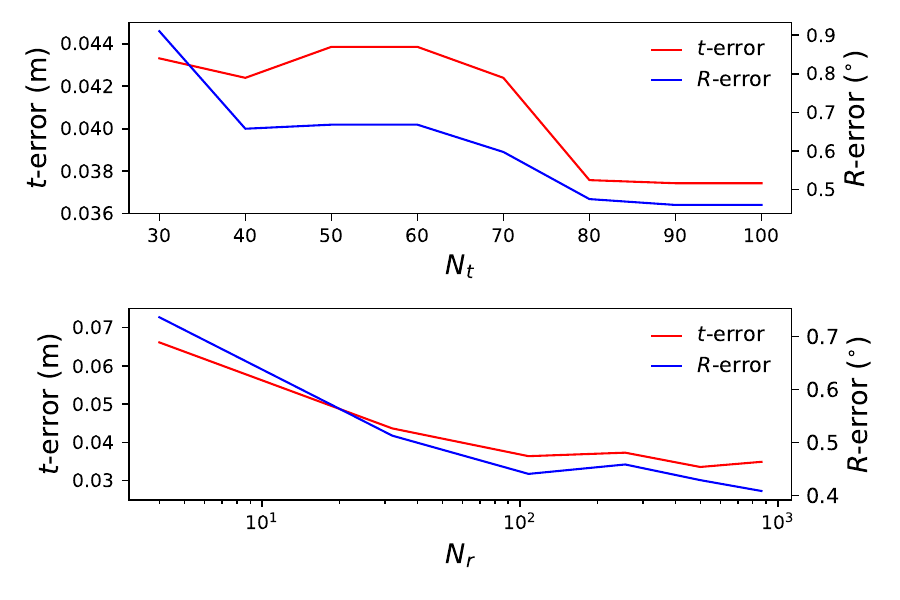}
    \vspace{-2em}
    \caption{Effects of $N_t$, $N_r$ on localization error.}
    \label{ablation_nt_nr}
    \end{subfigure}
    \newline
    \begin{subfigure}[t]{1.0\linewidth}
    \medskip
    \includegraphics[width=0.62\linewidth, valign=t]{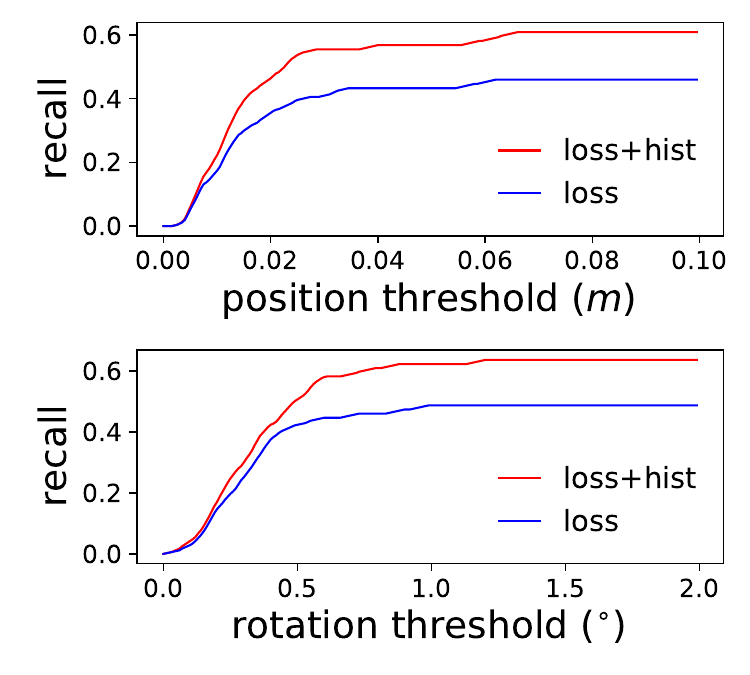}
    \includegraphics[width=0.37\linewidth, valign=t]{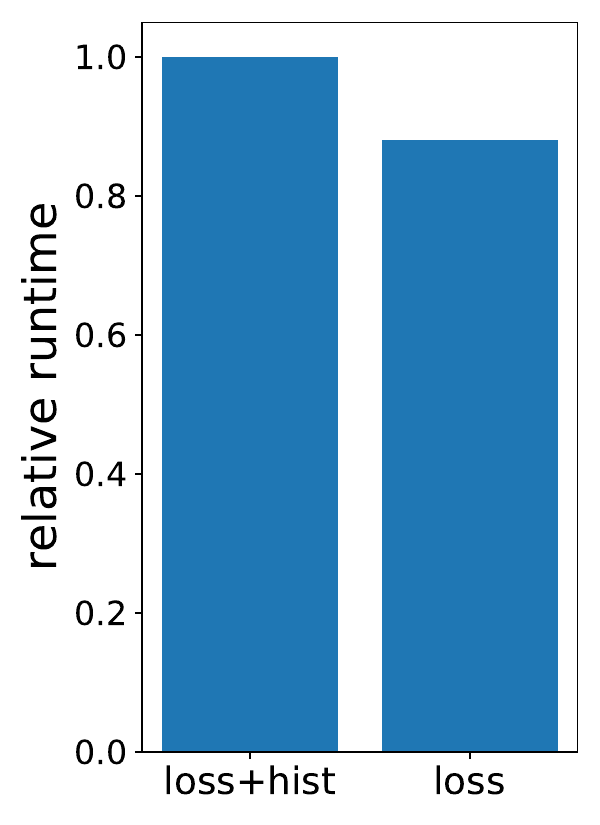}
    \caption{Comparison of two initialization schemes.}
    \label{ablation_initialization}
    \end{subfigure}
\vspace{-0.5em}
\caption{Ablation study on the initialization pipeline.}
\label{ablation}
\vspace{-0.5em}
\end{figure}

\paragraph{Gravity Direction}
If the gravity direction is known, the number of initial positions is significantly reduced and PICCOLO can perform highly accurate localization as shown in Table~\ref{table:photometric}.
Knowing the gravity direction is a reasonable assumption as many panoramic images or 3D scan datasets~\cite{stanford2d3d, fukuoka} contain the information.
In practice, one can easily infer the gravity direction of omnidirectional cameras using integrated gyroscopes, and that of 3D maps with RANSAC~\cite{ransac}-based plane fitting. 
Nonetheless, PICCOLO stably performs without knowing the gravity direction as shown in Table~\ref{table:stanford_all} and~\ref{table:fukuoka}.

In case PICCOLO might be biased towards the gravity-aligned conventional data, we evaluate PICCOLO in flipped input images and arbitrarily rotated point clouds.
Under the same hyperparameter setup as Section~\ref{perf}, PICCOLO demonstrates consistent performance, as shown in Table~\ref{table:photometric}.
Such results imply that PICCOLO is amenable for novel scenes, and could be directly applied to a wide variety of non-standard inputs without training.

\paragraph{Initialization Pipeline}
We finally ablate various components of the initialization pipeline presented in Section~\ref{sec:init}.
The first main parameters to be examined are the number of initial points $N_t, N_r$ sampled from the range of possible transformations.
We evaluate the effect of different values of $N_t, N_r$ on auditoriums from Area 2 of the Stanford2D-3D-S dataset~\cite{stanford2d3d}.
As shown in Figure~\ref{ablation_nt_nr}, larger $N_t, N_r$ tend to improve the error values, but result in computational overhead.
An adequate set of $N_t, N_r$ should be chosen considering the trade-off.
We use $N_t=50, N_r=32$ for all our experiments.
This means we have about 1600 initial points to test, but the initialization finishes within a few seconds, thanks to the efficiency of sampling loss.
For runtime-critical applications, one may cache the projected point cloud coordinates at each candidate starting pose once for each scene and use it afterward.
This would significantly reduce the time spent on initialization. 

We further examine the efficacy of our two-stage initialization scheme.
Recall the two-stage initialization in Section~\ref{sec:init} first selects $K_1$ candidate locations using \emph{loss values} followed by filtration to $K_2$ candidates using \emph{color histograms}.
We compare the performance of PICCOLO selecting $K_2$ initial poses from $N_t\times N_r$ candidates using (i) loss only, and (ii) the two-stage method presented in Section~\ref{sec:init}.
All rooms in Area 3 of the Stanford2D-3D-S dataset are selected for evaluation with $N_t=50, N_r=32, K_1=50, K_2=6$.
We display the results in Figure~\ref{ablation_initialization}.
Our two-stage initialization enables a significant performance boost with only a small increase in runtime.

\afterpage{\clearpage}
\begin{figure*}[p]
\begin{center}

\begin{subfigure}[b]{\textwidth}
\makebox[\textwidth][c]{
    \includegraphics[width=0.19\linewidth]{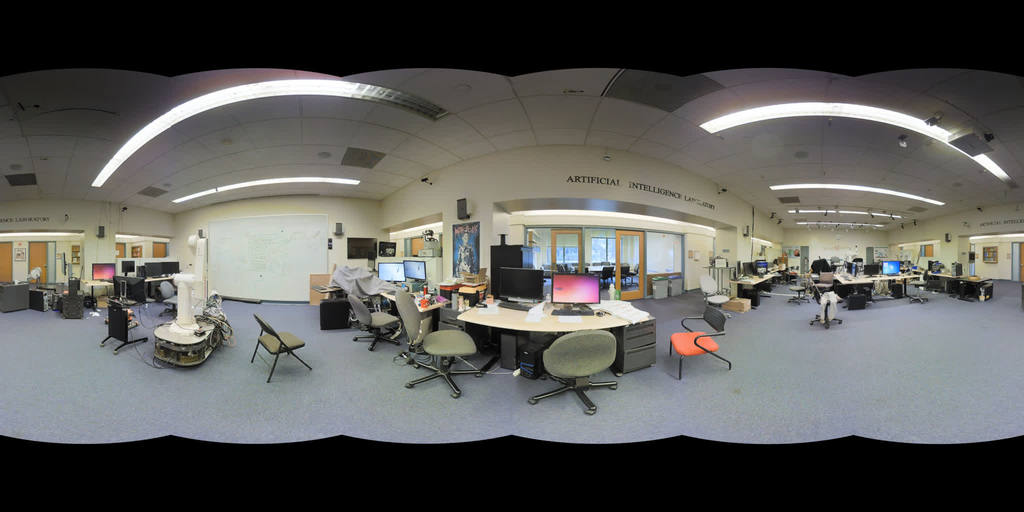}
    \includegraphics[width=0.19\linewidth]{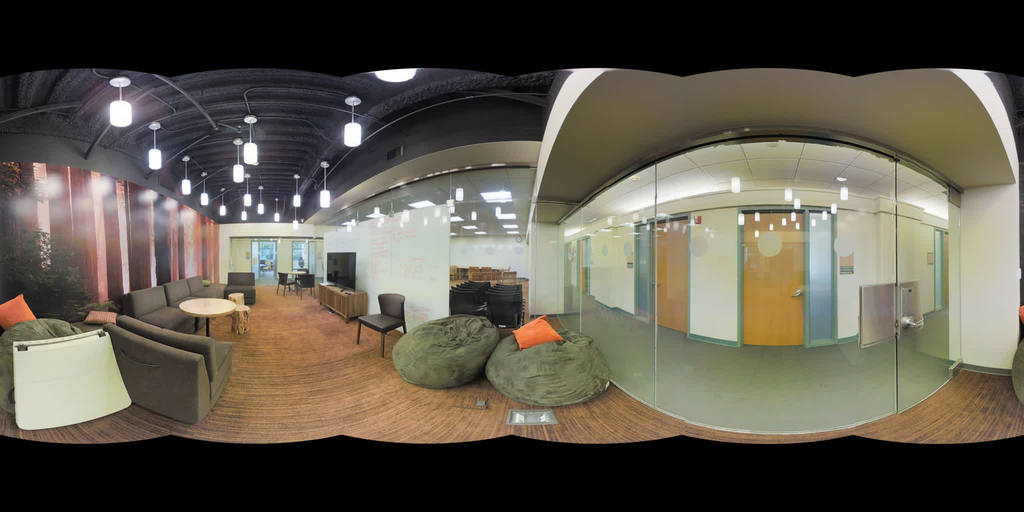}
    \includegraphics[width=0.19\linewidth]{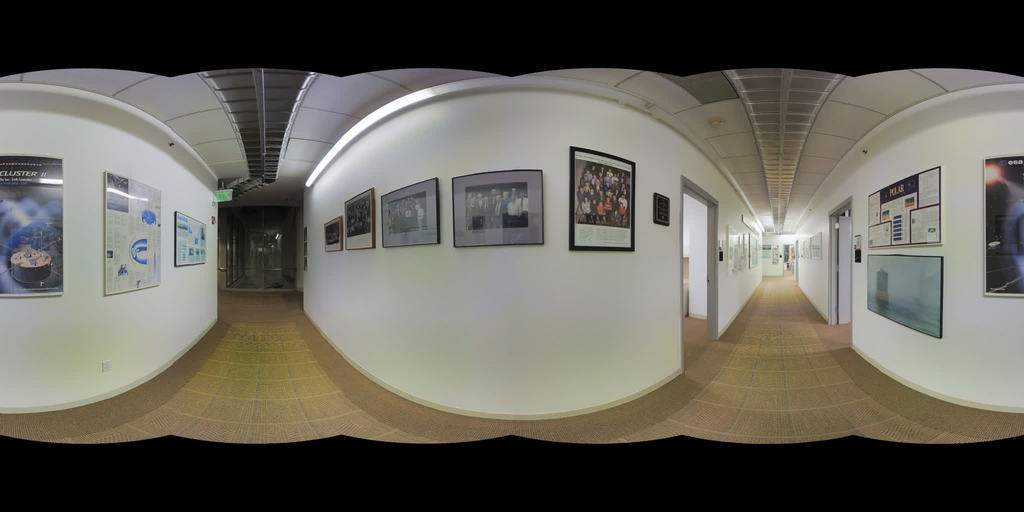}
    \includegraphics[width=0.19\linewidth]{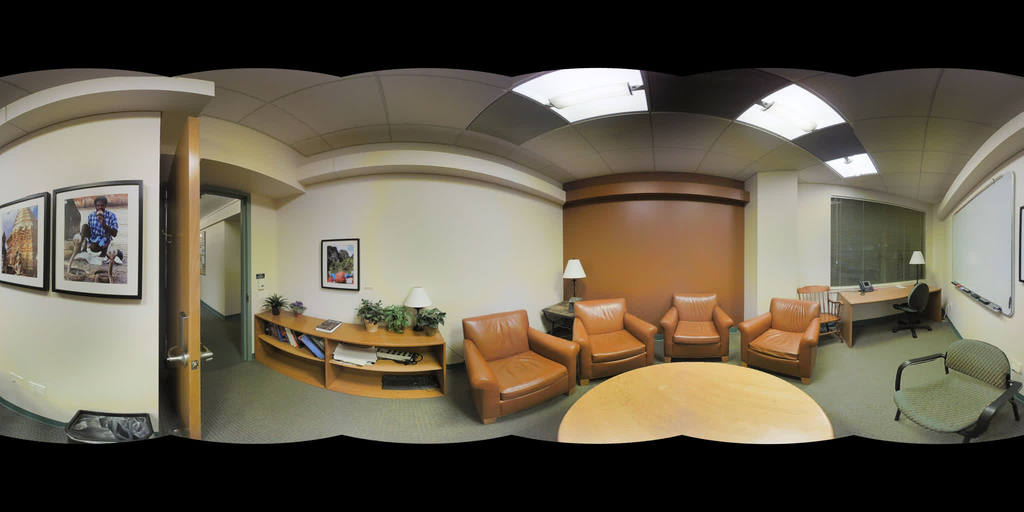}
    \includegraphics[width=0.19\linewidth]{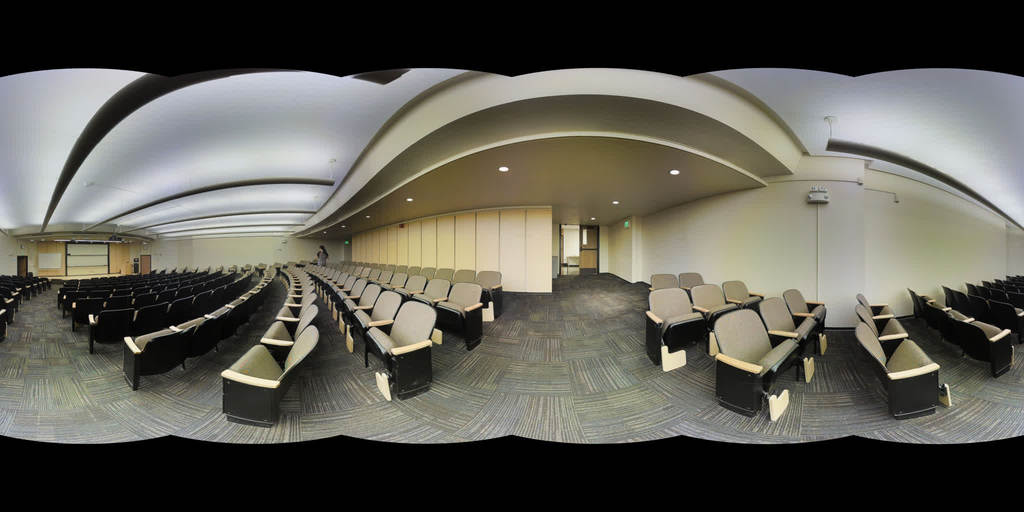}
    }
\makebox[\textwidth][c]{
    \includegraphics[width=0.19\linewidth]{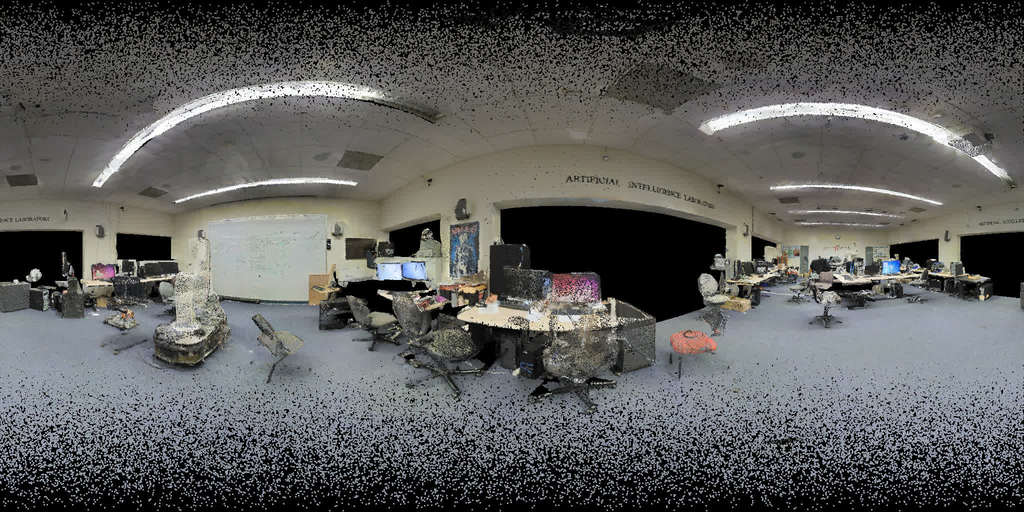}
    \includegraphics[width=0.19\linewidth]{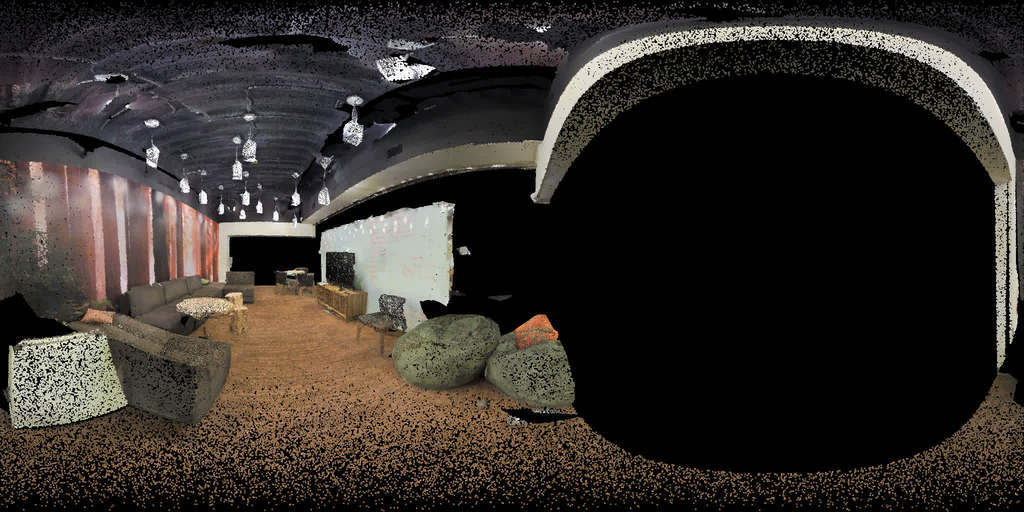}
    \includegraphics[width=0.19\linewidth]{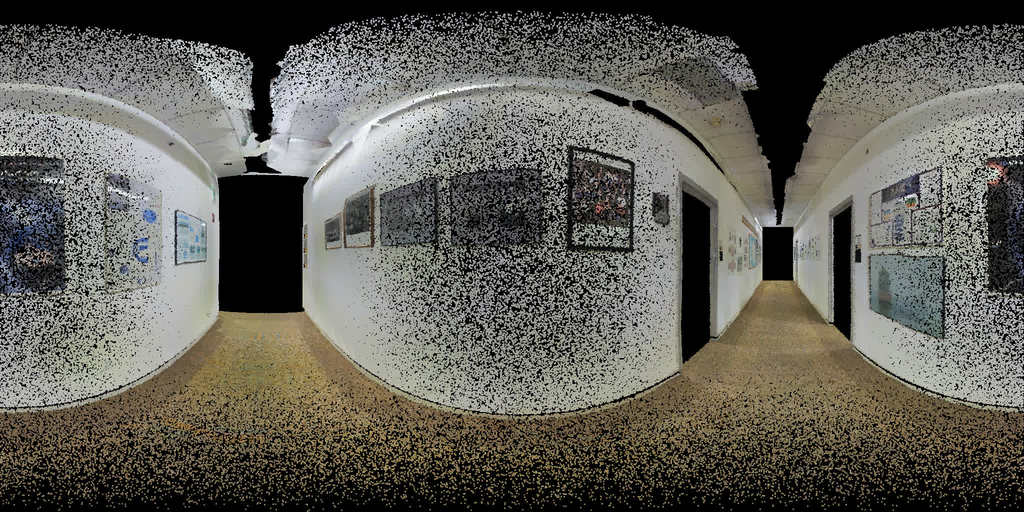}
    \includegraphics[width=0.19\linewidth]{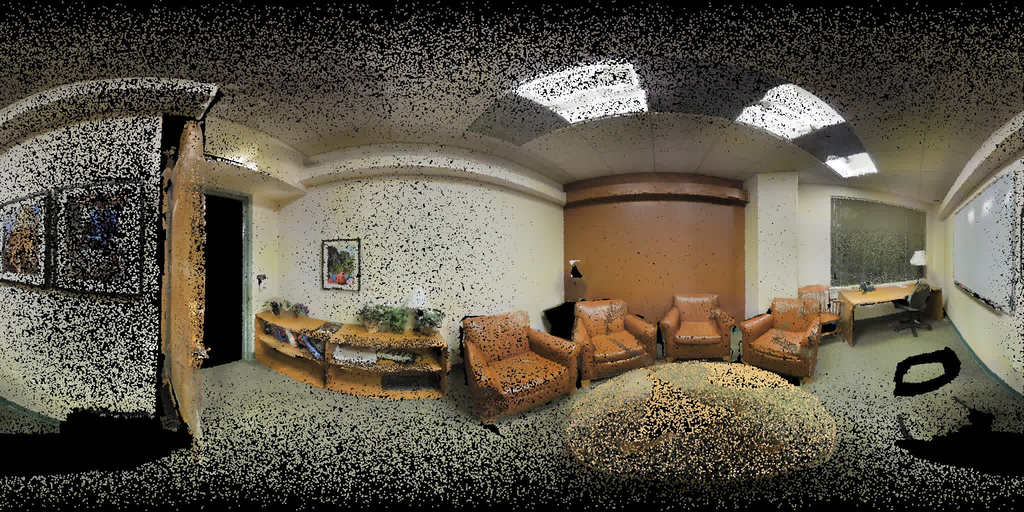}
    \includegraphics[width=0.19\linewidth]{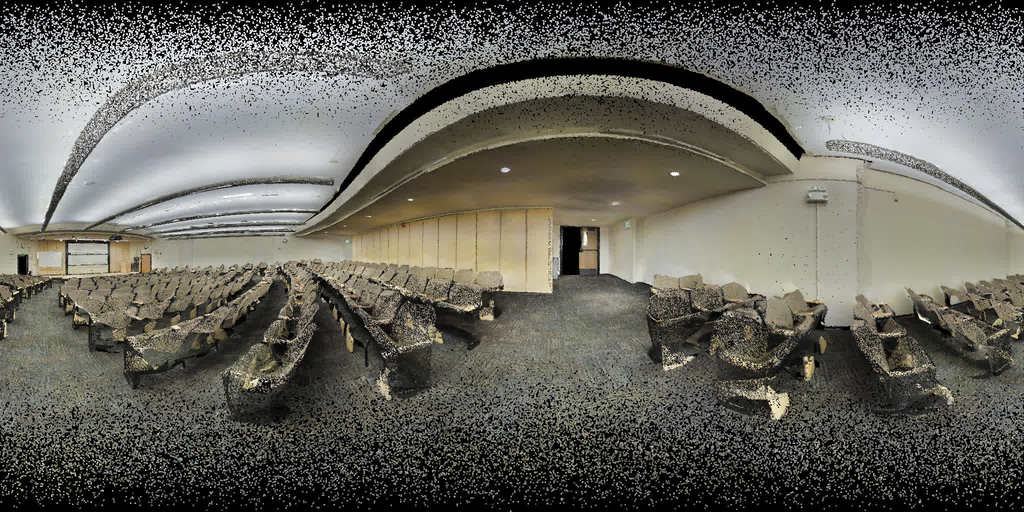}
    }
    \caption{Stanford2D-3D-S Indoor Localization}
\end{subfigure}

\begin{subfigure}[b]{\textwidth}
\makebox[\textwidth][c]{
    \includegraphics[width=0.24\linewidth]{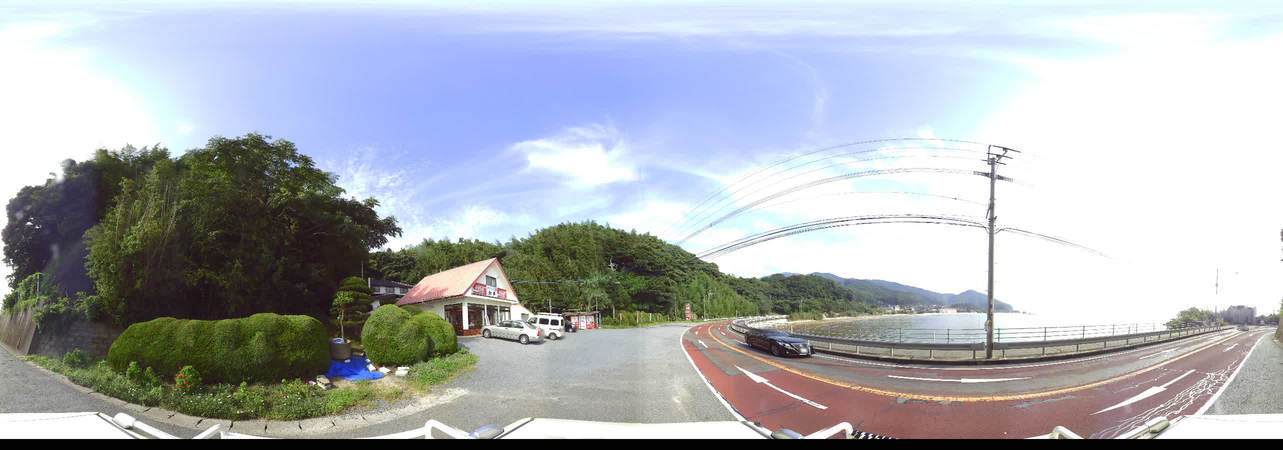}
    \includegraphics[width=0.24\linewidth]{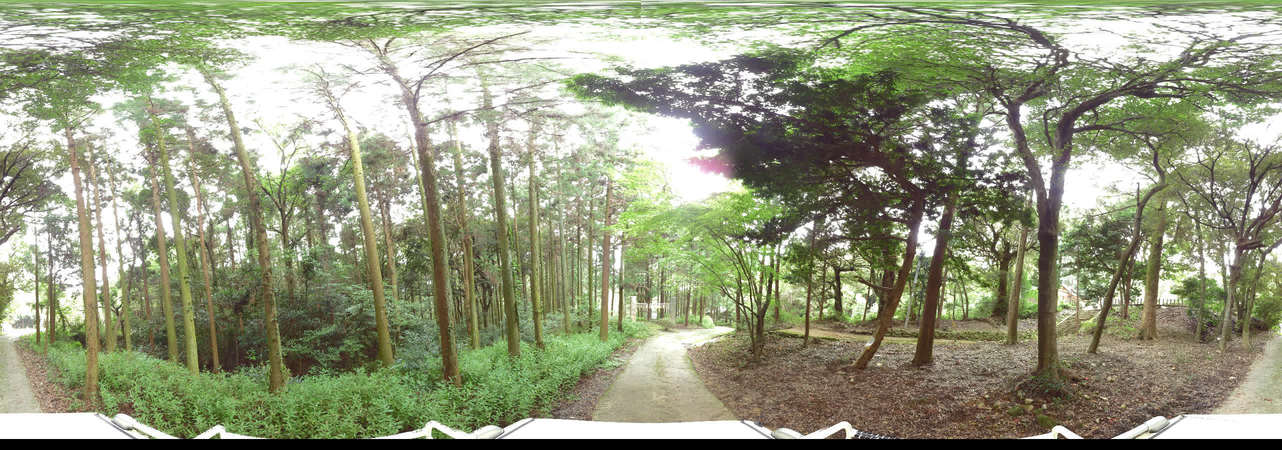}
    \includegraphics[width=0.24\linewidth]{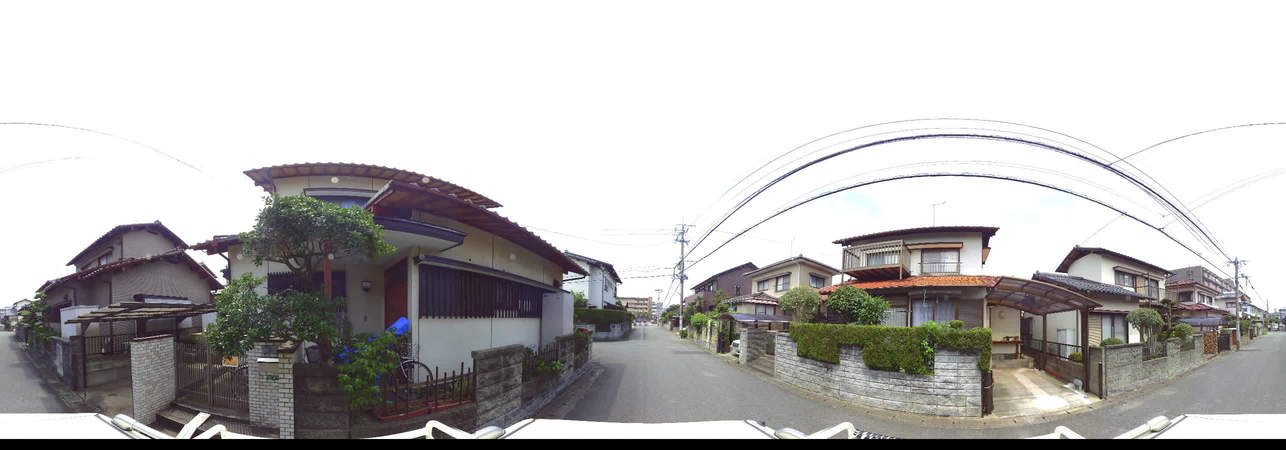}
    \includegraphics[width=0.24\linewidth]{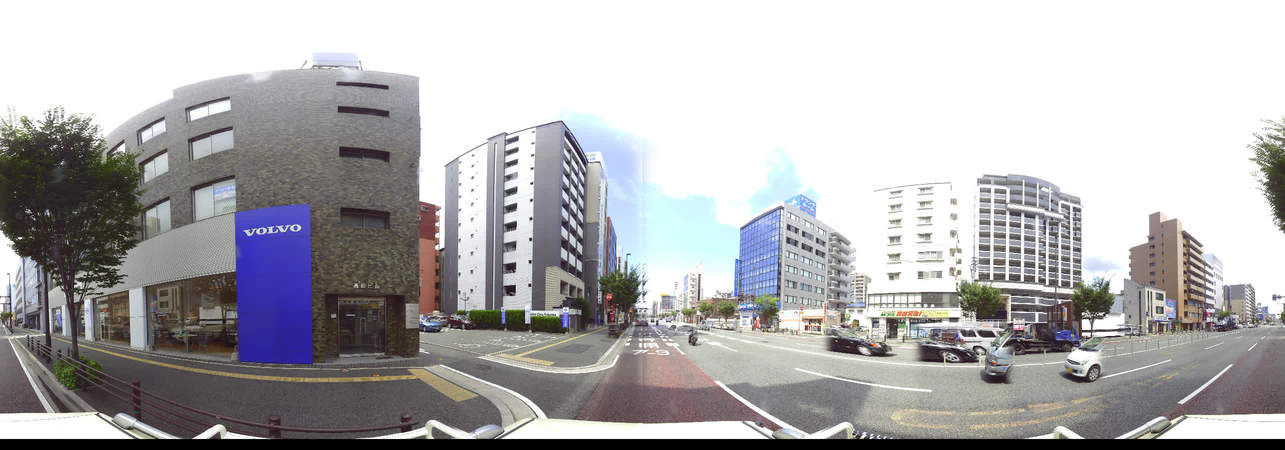}
    }
\makebox[\textwidth][c]{
    \includegraphics[width=0.24\linewidth]{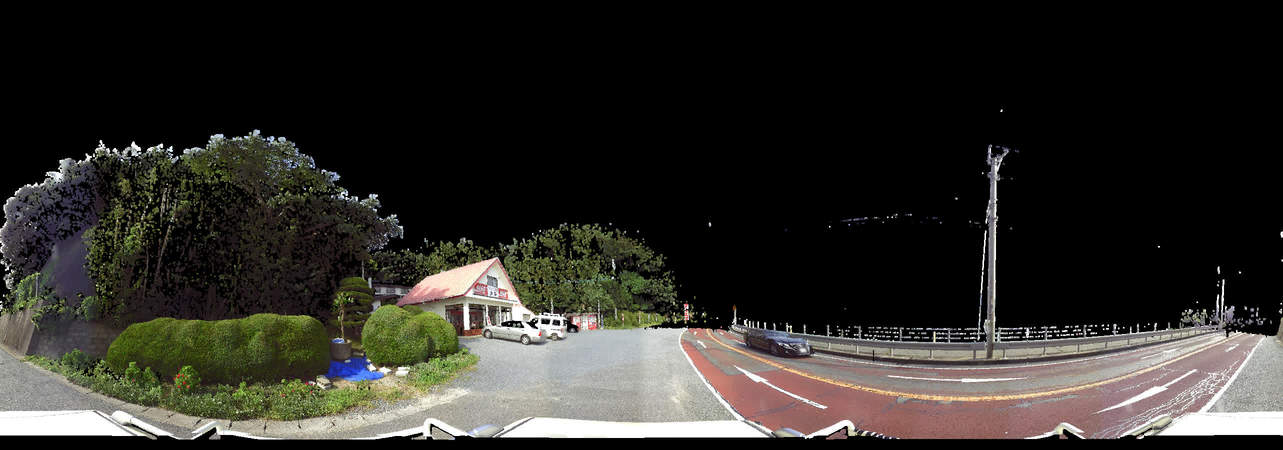}
    \includegraphics[width=0.24\linewidth]{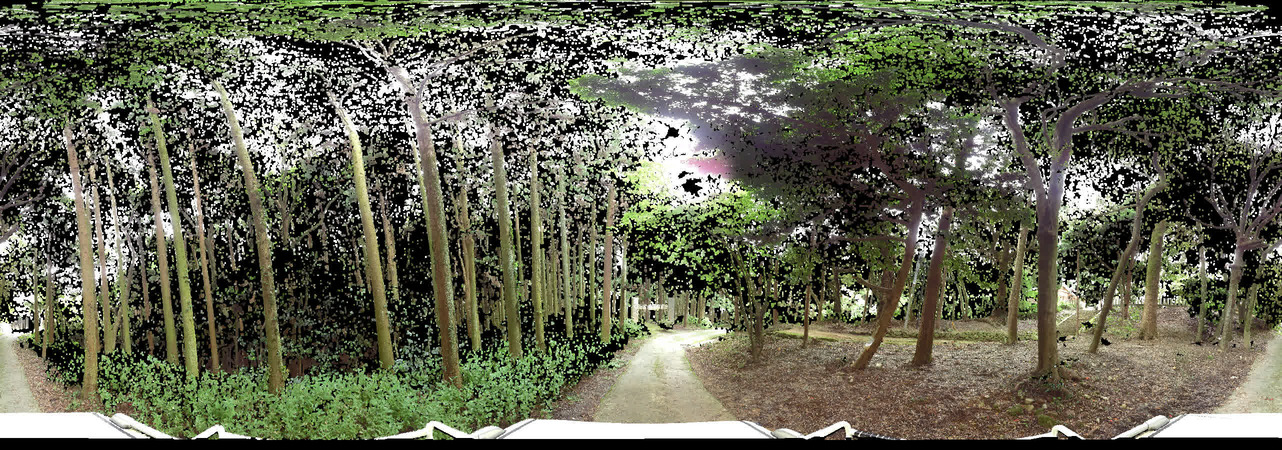}
    \includegraphics[width=0.24\linewidth]{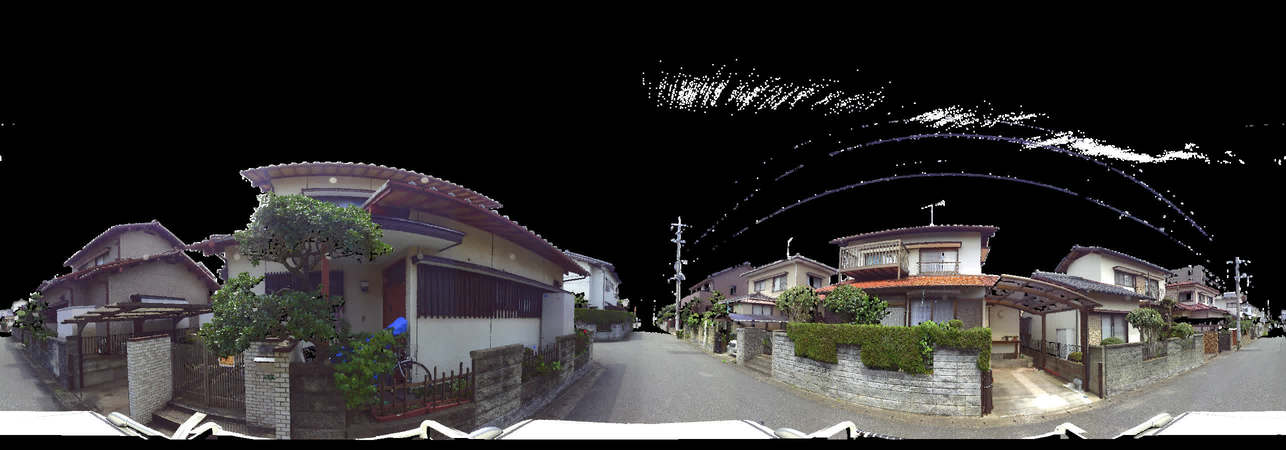}
    \includegraphics[width=0.24\linewidth]{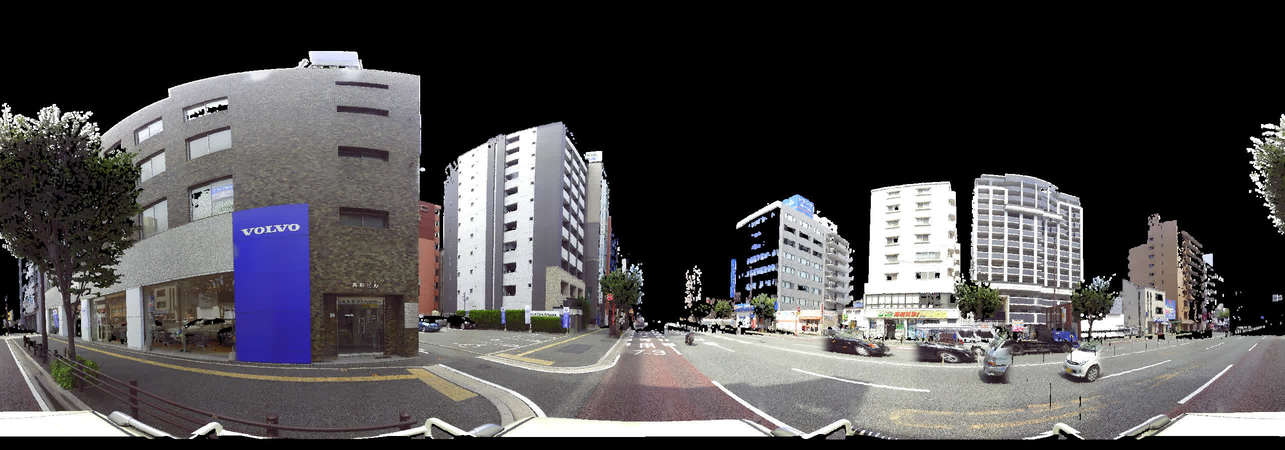}
    }
    \caption{MPO Outdoor Localization}
\end{subfigure}

\begin{subfigure}[t]{0.42\textwidth}
\makebox[\textwidth][c]{
    \includegraphics[width=0.5\textwidth]{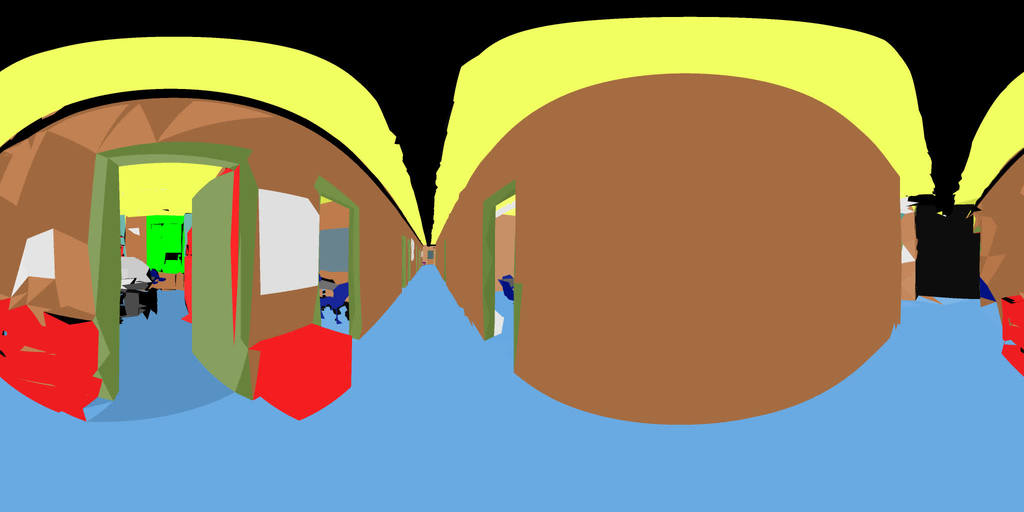}
    \includegraphics[width=0.5\textwidth]{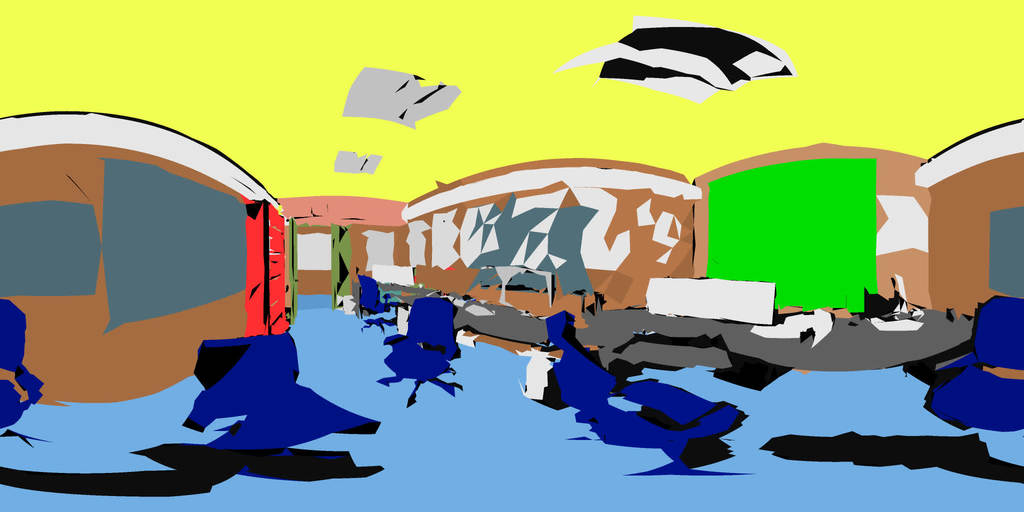}
}
\makebox[\textwidth][c]{
    \includegraphics[width=0.5\textwidth]{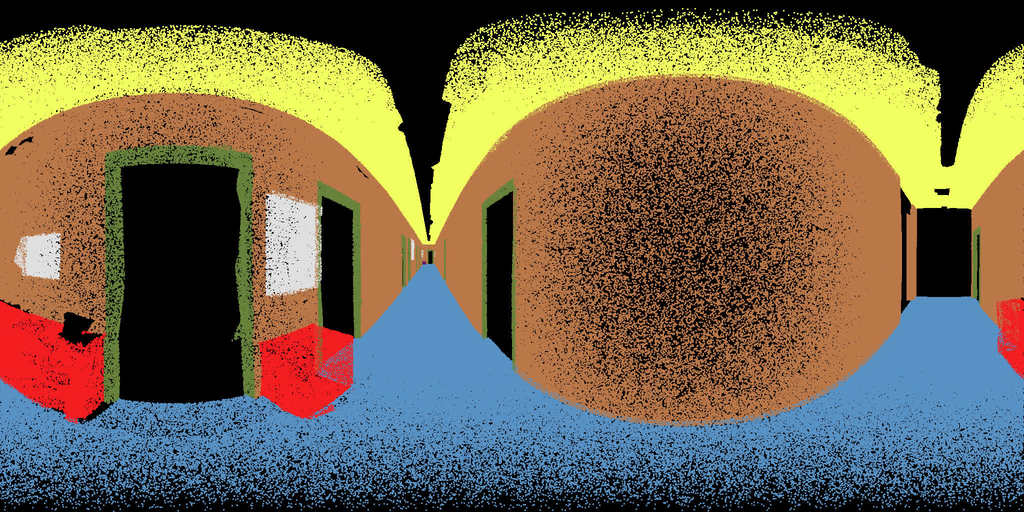}
    \includegraphics[width=0.5\textwidth]{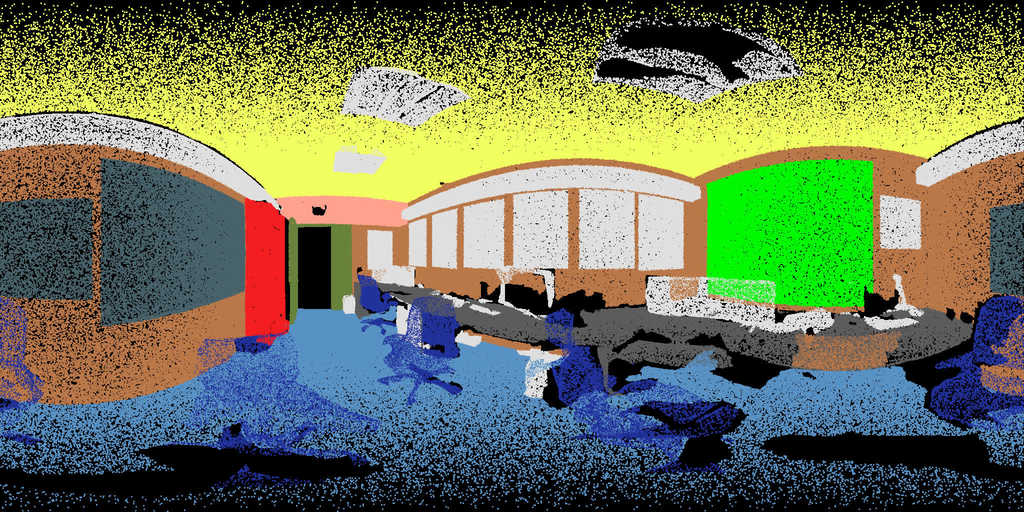}
}
\caption{Stanford2D-3D-S Semantic Input}
\end{subfigure}
\qquad
\begin{subfigure}[t]{0.42\textwidth}
\makebox[\textwidth][c]{
    \includegraphics[width=0.5\textwidth]{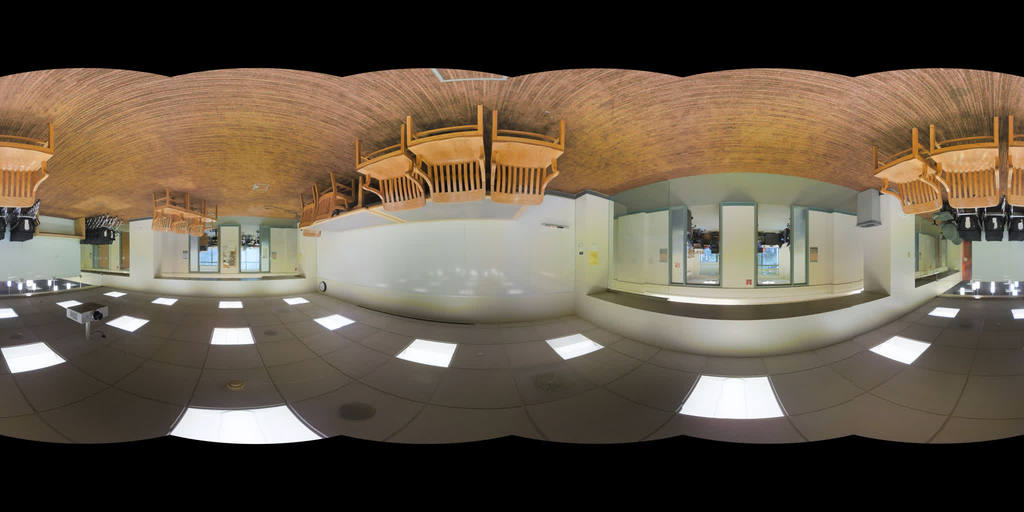}
    \includegraphics[width=0.5\textwidth]{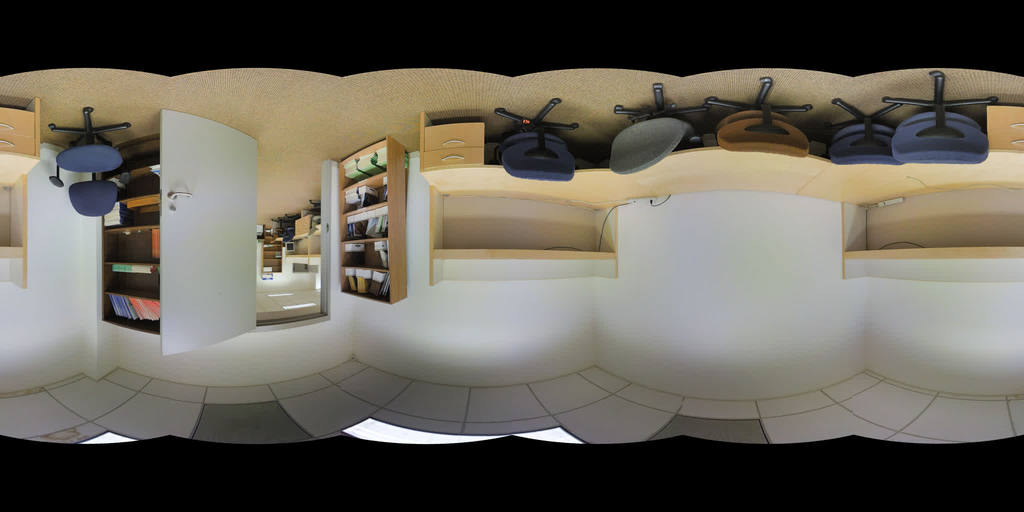}
}
\makebox[\textwidth][c]{
    \includegraphics[width=0.5\textwidth]{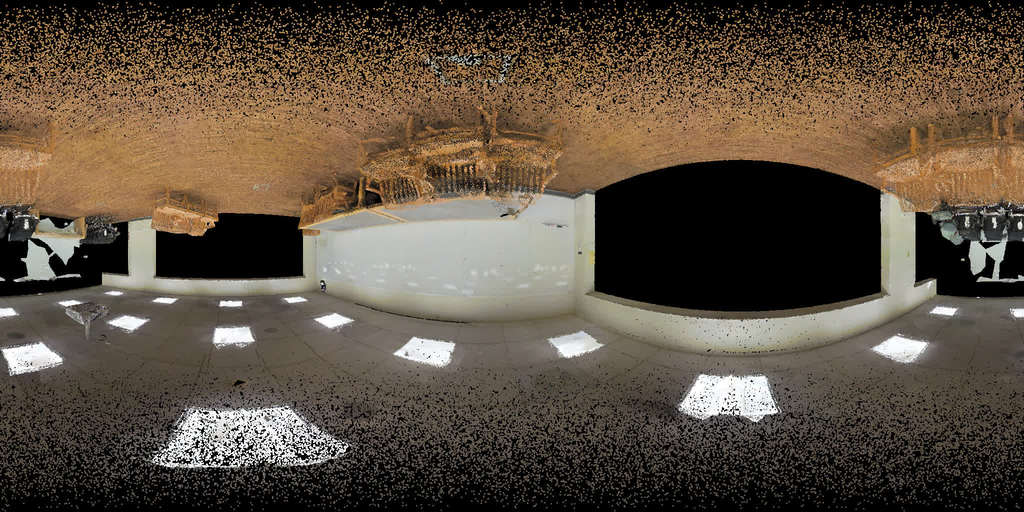}
    \includegraphics[width=0.5\textwidth]{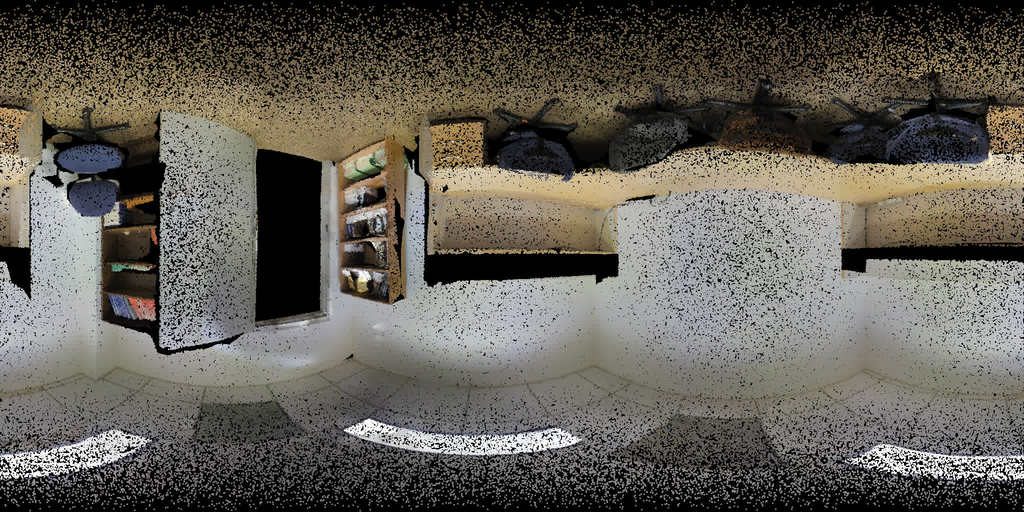}
}
\caption{Stanford2D-3D-S Flipped Input}
\end{subfigure}


\begin{subfigure}[t]{0.47\textwidth}
\makebox[\textwidth][c]{
\includegraphics[width=0.5\textwidth]{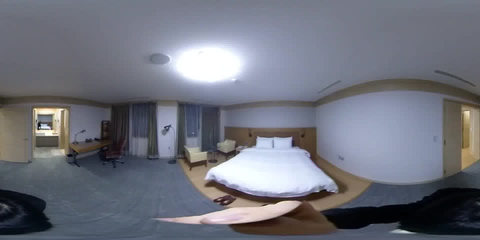}
    \includegraphics[width=0.5\textwidth]{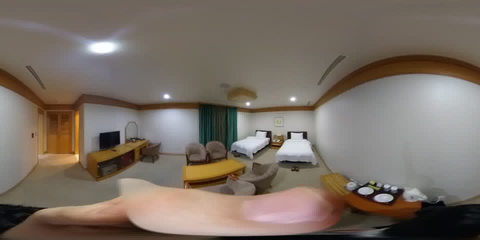}
}
\makebox[\textwidth][c]{
    \includegraphics[width=0.5\textwidth]{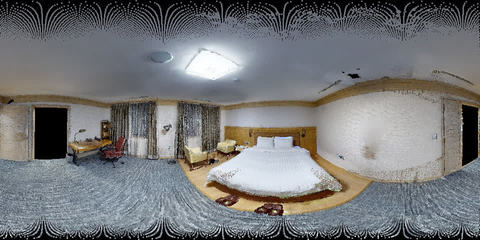}
    \includegraphics[width=0.5\textwidth]{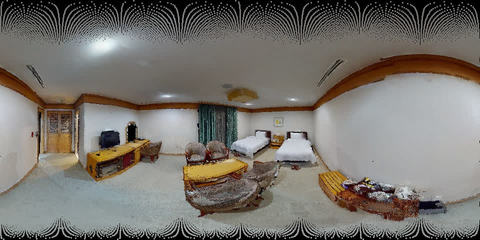}
}
\caption{OmniScenes Handheld}
\end{subfigure}
\qquad
\begin{subfigure}[t]{0.47\textwidth}
\makebox[\textwidth][c]{
    \includegraphics[width=0.5\textwidth]{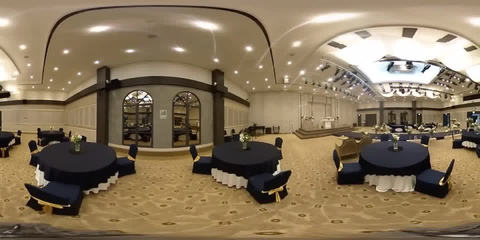}
    \includegraphics[width=0.5\textwidth]{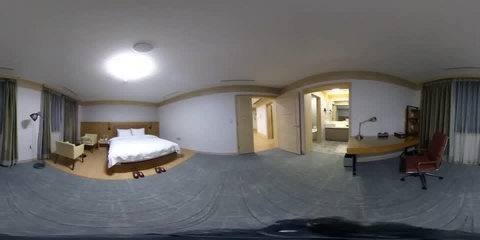}
}
\makebox[\textwidth][c]{
    \includegraphics[width=0.5\textwidth]{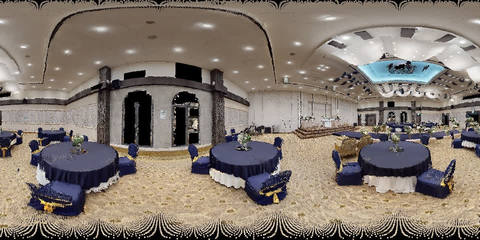}
    \includegraphics[width=0.5\textwidth]{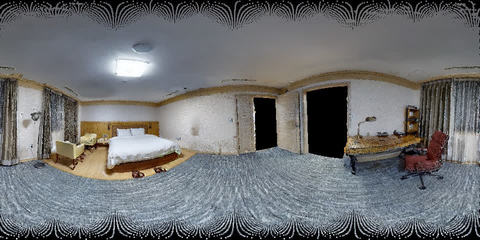}
}
\caption{OmniScenes Robot-Mounted}
\end{subfigure}
\begin{subfigure}[t]{0.47\textwidth}
\makebox[\textwidth][c]{
    \includegraphics[width=0.5\textwidth]{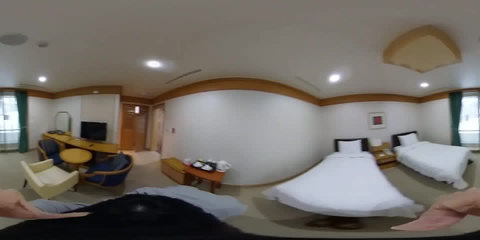}
    \includegraphics[width=0.5\textwidth]{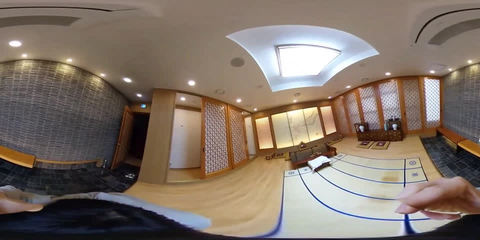}
}
\makebox[\textwidth][c]{
    \includegraphics[width=0.5\textwidth]{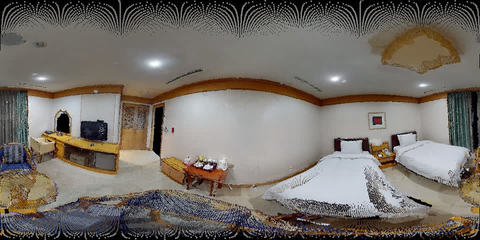}
    \includegraphics[width=0.5\textwidth]{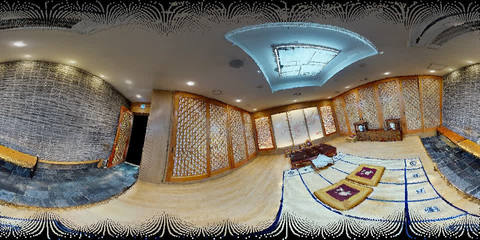}
}
\caption{OmniScenes Handheld with Scene Change}
\end{subfigure}
\qquad
\begin{subfigure}[t]{0.47\textwidth}
\makebox[\textwidth][c]{
    \includegraphics[width=0.5\textwidth]{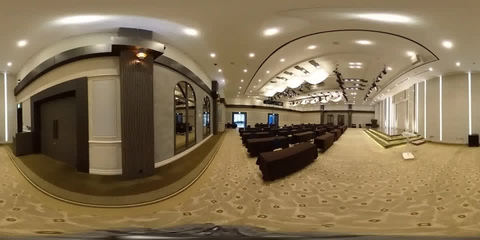}
    \includegraphics[width=0.5\textwidth]{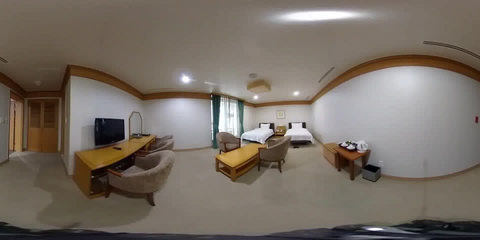}
}
\makebox[\textwidth][c]{
    \includegraphics[width=0.5\textwidth]{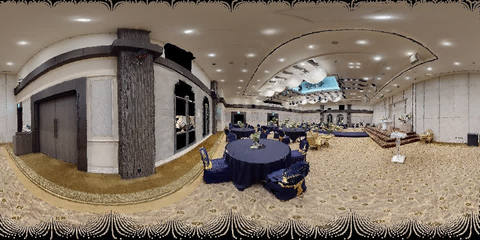}
    \includegraphics[width=0.5\textwidth]{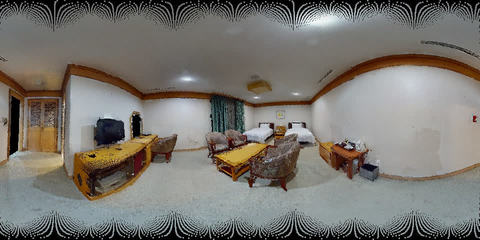}
}
\caption{OmniScenes Robot-Mounted with Scene Change}
\end{subfigure}
\end{center}
\vspace{-1.5em}
\caption{Additional qualitative results of PICCOLO with various input settings. We display the input query image (top) and the projected point cloud under the estimated camera pose (bottom).}
\label{qualitative_results}
\end{figure*}

\section{Conclusion}
In this paper, we present PICCOLO, a simple, efficient algorithm for omnidirectional localization.
We introduce sampling loss, which enforces each point in the 3D point cloud to correctly \emph{sample} from the query image.
Sampling loss is clearly beneficial for omnidirectional localization, as it is a \emph{point cloud-centric} formulation, free from spherical distortion, and computationally efficient.
In experiments conducted on various indoor and outdoor environments, PICCOLO outperforms existing algorithms by a significant margin.
Furthermore, when evaluated on our newly proposed dataset, OmniScenes, PICCOLO shows competent performance even amidst diverse camera motion and scene changes.
We expect PICCOLO to be applied in a wide variety of virtual reality / robotics applications where omnidirectional cameras are present.
\paragraph{Acknowledgments} This research was supported by the National Research Foundation of Korea (NRF) grant funded by the Korea government (MSIT) (No. 2020R1C1C1008195), the National Convergence Research of Scientific Challenges through the National Research Foundation of Korea (NRF) funded by Ministry of Science and ICT (NRF2020M3F7A1094300), and the BK21 FOUR program of the Education and Research Program for Future ICT Pioneers, Seoul National University in 2021.

\appendix
\renewcommand\thetable{\thesection.\arabic{table}}    
\setcounter{table}{0}
\renewcommand\thefigure{\thesection.\arabic{figure}}    
\setcounter{figure}{0}
\def\UrlBreaks{\do\/\do-}
\iccvfinalcopy
\def\iccvPaperID{3864} 
\def\confYear{ICCV 2021}

\section{Additional Implementation Details}
In this section, we provide several additional details in PICCOLO that are not provided in the original paper.
The majority of the description of PICCOLO is provided in Section 3.

\subsection{Gradient Step Size Scheduling}
To foster convergence, we adaptively decay the gradient step size $\alpha$ (line 7 of Algorithm 1) similarly to learning rate scheduling widely used in neural network training.
In our experiments, we decay the step size by a factor of 0.8 if the loss function value does not decrease for 5 consecutive iterations.

\section{Hyperparameter Setup}
We report the hyperparameter setups of PICCOLO, where the configurations vary only depending on the providence of gravity direction.
As shown in Algorithm 1, we first sample $N_r \times N_t$ initial poses, out of which we select $K_1$ poses through loss value-based filtering, which are consecutively reduced into $K_2$ poses using color histogram intersection. 
Then we run gradient descent for $K_2$ starting points for $N_{iter}$ steps.
The quantitative results are reported in Section 4.1.

\subsection{Unknown Gravity Direction}
For inputs where the gravity direction is unknown, we use $N_r=32, N_t=50, N_{iter}=100, K_1=50, K_2=6$.
Such a setup is used in Table 1, 2, 3, and 5 (excluding `gravity direction').
This setup shows effective performance in both indoor / outdoor datasets (Stanford2D-3D-S~\cite{stanford2d3d}, MPO~\cite{fukuoka}), and generalizes to arbitrary point cloud rotations and flipped images.
The capacity of PICCOLO to generalize in such diverse inputs under a fixed hyperparameter configuration alludes to its potential as an off-the-shelf omnidirectional localization algorithm.

\subsection{Known Gravity Direction}
We use the following setup for inputs where the gravity direction is known: $N_r=8, N_t=50, N_{iter}=100, K_1=50, K_2=6$.
Such a setup is used in Table 4 and 5 (`gravity direction').
Since the gravity direction is known, the search space can be dramatically reduced, and PICCOLO can perform highly accurate localization, as shown in Table 4 and 5.

\setcounter{table}{0}
\setcounter{figure}{0}
\section{Dataset Details}
Here we report minor experimental details about the datasets used in Section 4.
For all datasets, we remove panoramas where the ground truth camera position is outside the point cloud's bounding box.

\subsection{Stanford2D-3D-S}
For GOSMA~\cite{gosma}, if the room size exceeds a certain limit, segmentation fault occurs and the algorithm terminates.
We exclude those cases when computing the accuracy.
We additionally describe the 33 rooms used for creating Table 2. 
These rooms are chosen using the criterion of Campbell \etal~\cite{gosma}: (i) distance to the closest point is greater than 50 cm, (ii) number of labeled pixels should be greater than 2000, (iii) number of 3D points should be greater than 2000.

\subsection{MPO}
A peculiar aspect of the MPO dataset is that all the ground truth values are the same: they are fixed to $R^*=I, t^*=\mathbf{0}$.
Therefore, we apply random rotation and translation to the point cloud to make the dataset more challenging.
For rotation, we randomly rotate the point cloud along the z-axis by an angle $\theta \sim \mathcal{U}(0, \pi)$, where $\mathcal{U}(\cdot)$ denotes the uniform distribution.
For translation, we randomly apply translations along the $x, y, z$ directions, where $x, y, z, \sim \mathcal{U}(0, 3)$. Note that the units for $x, y$, and $z$ are in meters.

\subsection{OmniScenes}
In this section, we report the acquisition process of our newly proposed OmniScenes dataset.
OmniScenes dataset is composed of 3D scans of 8 scenes accompanied by omnidirectional images.
3D scans are collected using the Matterport 3D scanner~\cite{matterport_scan}.
The corresponding omnidirectional images are collected with the Ricoh Theta $360^\circ$ camera~\cite{Ricoh} under two scenarios: handheld and mobile robot-mounted.
For the handheld case, we have the capturer to take $360^\circ$ videos while walking around the eight scenes.
For the mobile robot-mounted case, we use a TurtleBot3 Waffle~\cite{turtle} that is manually controlled.
To obtain 6DoF pose annotations for each omnidirectional image, we apply SfM (Structure from Motion), as in Kendall~\etal~\cite{posenet}.
COLMAP~\cite{colmap_1, colmap_2} and OpenMVG~\cite{openmvg} are used to obtain dense SfM reconstructions, which are then manually aligned with the Matterport 3D scans, similar to Valentin~\etal~\cite{energy_landscape}.
The aligned SfM model contains omnidirectional camera poses with respect to the Matterport 3D scans.
We further remove invalid pose estimates from SfM through manual inspection.
The resulting dataset, OmniScenes, contains a wide variety of pose-annotated omnidirectional images.
The statistical properties of our dataset are displayed in Table~\ref{table:omniscenes_supp}.

\begin{table}[t]
    \centering
    \newcolumntype{C}{>{\centering\arraybackslash}X}
    \renewcommand*{\arraystretch}{0.95}
    {\small
    \begin{tabularx}{\columnwidth}{lC|CC}
        \toprule
        Scenario & Scene Change & \# of Images & \# of Scenes\\
        \midrule
        Handheld & $\bigtimes$ & 1451 & 8 \\
        Robot & $\bigtimes$ & 1129 & 8 \\
        \midrule
        Handheld & $\bigcirc$ & 698 & 8 \\
        Robot & $\bigcirc$ & 1132 & 8 \\
        \bottomrule
    \end{tabularx}
    }
    \caption{Statistical Properties of the OmniScenes dataset.}
    \label{table:omniscenes_supp}
\end{table}

\setcounter{table}{0}
\setcounter{figure}{0}

\section{Qualitative Comparison with GOSMA~\cite{gosma}}
\begin{figure}[]
\makebox[\columnwidth][c]{
    \includegraphics[width=0.45\linewidth]{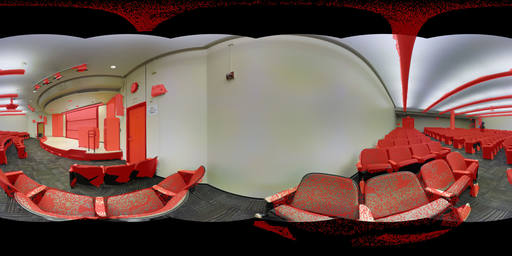}
    \includegraphics[width=0.45\linewidth]{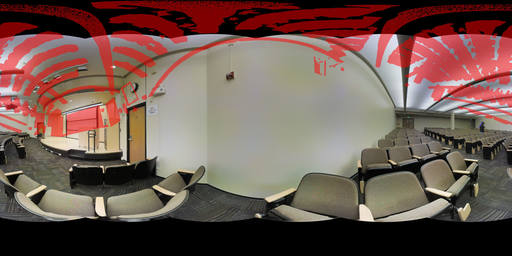}
}
\makebox[\columnwidth][c]{
    \includegraphics[width=0.45\linewidth]{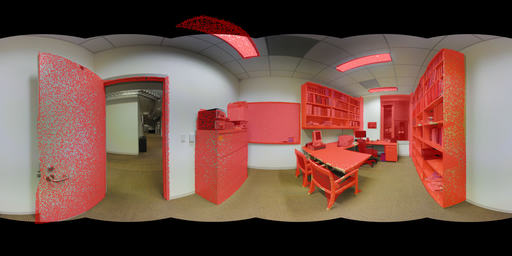}
    \includegraphics[width=0.45\linewidth]{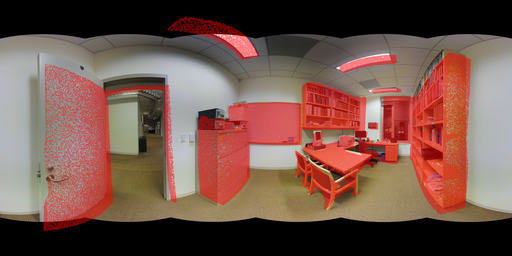}
}

\makebox[\columnwidth][c]{
    \includegraphics[width=0.45\linewidth]{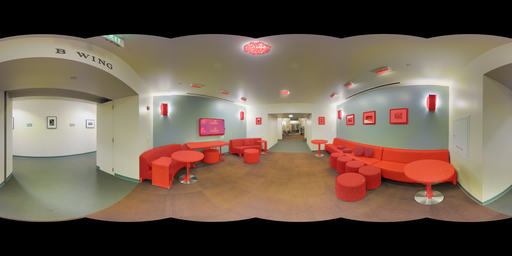}
    \includegraphics[width=0.45\linewidth]{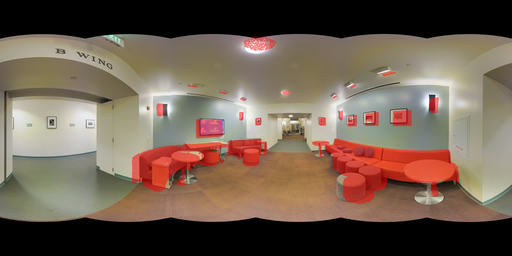}
}

\makebox[\columnwidth][c]{
    \makebox[0.45\columnwidth][c]{
        \textit{PICCOLO}
    }
    \makebox[0.45\columnwidth][c]{
        \textit{GOSMA}
    }
}
\caption{Qualitative results of PICCOLO and GOSMA~\cite{gosma} in Stanford2D-3D-S dataset~\cite{stanford2d3d}. Projected point cloud coordinates are overlayed in red.}
\label{comparison}
\end{figure}

We make a qualitative comparison of PICCOLO and GOSMA in Figure~\ref{comparison}.
While GOSMA is successful in small rooms such as offices, it often fails in large areas such as auditoriums.
PICCOLO is capable of performing stable omnidirectional localization under diverse challenging environments.
Further, recall that semantic labels were given as input, to ensure fair comparison with GOSMA~\cite{gosma}.
This indicates that PICCOLO can function seamlessly with any other point-wise information.

\setcounter{table}{0}
\setcounter{figure}{0}
\section{Runtime Analysis}
\begin{table}[t]
    \centering
    \newcolumntype{C}{>{\centering\arraybackslash}X}
    \renewcommand*{\arraystretch}{0.95}
    {\small
    \begin{tabularx}{\columnwidth}{@{}C|CC@{}}
        \toprule
        \# of Points & Initialization& Gradient Descent\\
        \midrule
        $10^4$ & 1.77 & 0.01 \\
        \midrule
        $10^5$ & 1.90 & 0.01 \\
        \midrule
        $10^6$ & 3.74 & 0.02 \\
        \bottomrule
    \end{tabularx}
    }
    
    \caption{Runtime analysis of PICCOLO. All runtime statistics are reported in seconds.}
    \label{table:runtime}
\end{table}

In this section, we examine the runtime properties of PICCOLO.
Experiments are conducted with a single RTX 2080 GPU and an Intel Core  i7-9700 3.00GHz CPU.
We report the amount of time it takes for initialization and gradient descent in Table~\ref{table:runtime}, with a varying number of points in the input point cloud.
We use the same configuration used for unknown gravity direction: $N_r=32, N_t=50, N_{iter}=100, K_1=50, K_2=6$.

Table~\ref{table:runtime} shows that initialization (line 2, 3 of Algorithm 1) terminates within a few seconds, and gradient descent  (line 7 of Algorithm 1) finishes on the scale of milliseconds.
Note that both initialization and gradient descent are easily parallelizable.
Thus PICCOLO can directly benefit from the presence of multiple GPUs, similar to GOSMA~\cite{gosma} and GOPAC~\cite{gopac}.
Furthermore, while the number of points is increased by a factor of 10, the runtime only shows a modest increase.
Sampling loss scales seamlessly to large point clouds, as no costly operations such as visibility computation take place.

\section{Additional Ablation Study}
In this section, we perform additional ablation study on PICCOLO.
We examine the number of starting points on initialization runtime and the effect of point cloud density on pose error.
The results are displayed in Figure~\ref{ablation}.

\begin{figure}
\centering
    \begin{subfigure}[t]{1.0\linewidth}
    \centering
    \includegraphics[width=\linewidth]{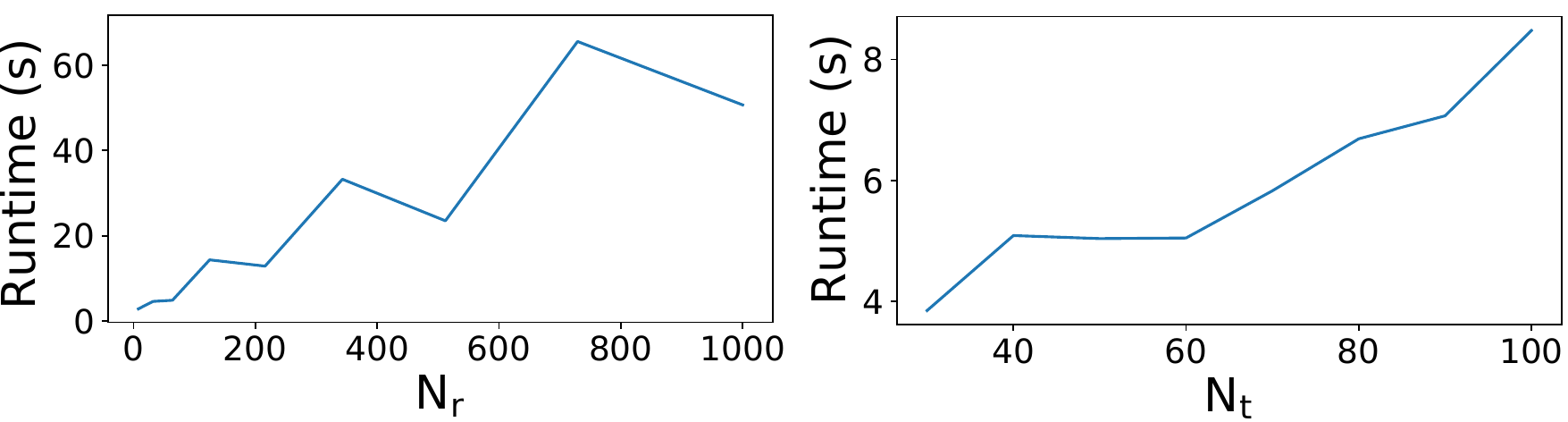}
    \caption{Effects of $N_t$, $N_r$ on initialization runtime.}
    \label{time}
    \end{subfigure}
    \newline
    \begin{subfigure}[t]{0.9\linewidth}
    \centering
    \includegraphics[width=\linewidth]{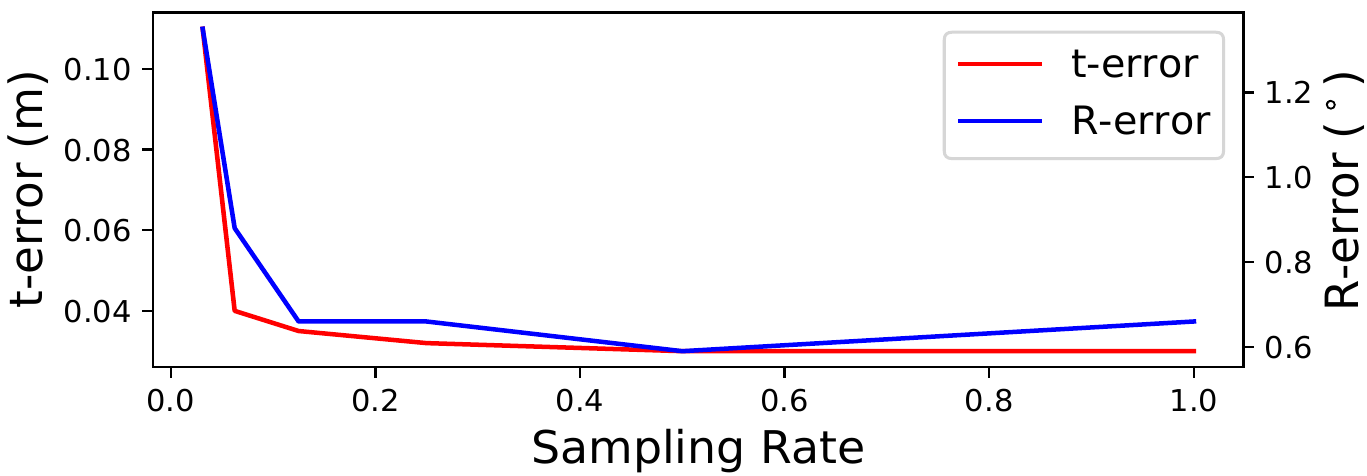}
    \caption{Effect of point cloud sampling rate on localization error.}
    \label{sp}
    \end{subfigure}
\caption{Additional ablation study on PICCOLO.}
\label{ablation}
\end{figure}

\paragraph{Number of Starting Points} PICCOLO is tested with varying numbers of translation and rotation starting points $N_t, N_r$ on offices from Area 2 of the Stanford2D-3D-S~\cite{stanford2d3d} dataset.
The median initialization runtime is reported with either $N_t$ or $N_r$ modified from the original setup.
Figure~\ref{time} indicates that while increasing $N_t, N_r$ leads to enhanced performance as shown in Figure 4, this also incurs longer runtime.
An appropriate $N_t, N_r$ that balances the trade-off should be chosen.
We use $N_t=50, N_r=32$ in all our experiments, which shows competent performance yet maintains fast runtime.
\paragraph{Point Cloud Density} PICCOLO is evaluated with varying point cloud sampling rates on the Stanford2D-3D-S~\cite{stanford2d3d} dataset.
As seen in Figure~\ref{sp}, PICCOLO is robust against point cloud density: pose estimation error is very small even when less than $5\%$ of the entire point cloud is used.

\setcounter{table}{0}
\setcounter{figure}{0}

\bibliographystyle{ieee}
\bibliography{omni_bib}

\begin{thebibliography}{10}\itemsep=-1pt

\bibitem{matterport_scan}
Matterport 3d: How long does it take to scan a property?
\newblock
  \url{https://support.matterport.com/hc/en-us/articles/229136307-How-long-does-it-take-to-scan-a-property-}.
\newblock Accessed: 2020-02-18.

\bibitem{Ricoh}
Ricoh theta, experience the world in 360.
\newblock \url{https://theta360.com/en/}.
\newblock Accessed: 2021-03-16.

\bibitem{turtle}
What is turtlebot?
\newblock
  \url{https://emanual.robotis.com/docs/en/platform/turtlebot3/overview/}.
\newblock Accessed: 2021-03-16.

\bibitem{stanford2d3d}
Iro Armeni, Sasha Sax, Amir~R Zamir, and Silvio Savarese.
\newblock Joint 2d-3d-semantic data for indoor scene understanding.
\newblock {\em arXiv preprint arXiv:1702.01105}, 2017.

\bibitem{dsac}
Eric Brachmann, Alexander Krull, Sebastian Nowozin, J. Shotton, F. Michel, S.
  Gumhold, and C. Rother.
\newblock Dsac — differentiable ransac for camera localization.
\newblock {\em 2017 IEEE Conference on Computer Vision and Pattern Recognition
  (CVPR)}, pages 2492--2500, 2017.

\bibitem{gopac}
Dylan Campbell, Lars Petersson, Laurent Kneip, and Hongdong Li.
\newblock Globally-optimal inlier set maximisation for camera pose and
  correspondence estimation.
\newblock {\em IEEE Transactions on Pattern Analysis and Machine Intelligence},
  page preprint, June 2018.

\bibitem{gosma}
Dylan Campbell, Lars Petersson, Laurent Kneip, Hongdong Li, and Stephen Gould.
\newblock The alignment of the spheres: Globally-optimal spherical mixture
  alignment for camera pose estimation.
\newblock In {\em Proceedings of the 2019 IEEE/CVF Conference on Computer
  Vision and Pattern Recognition (CVPR)}, page to appear, Long Beach, USA, June
  2019. IEEE.

\bibitem{spherenet}
Benjamin Coors, Alexandru~Paul Condurache, and Andreas Geiger.
\newblock Spherenet: Learning spherical representations for detection and
  classification in omnidirectional images.
\newblock In {\em Proceedings of the European Conference on Computer Vision
  (ECCV)}, September 2018.

\bibitem{loc_feat_2}
Gabriela Csurka and Martin Humenberger.
\newblock From handcrafted to deep local invariant features.
\newblock {\em CoRR}, abs/1807.10254, 2018.

\bibitem{bundlefusion}
Angela Dai, Matthias Nie{\ss}ner, Michael Zoll{\"o}fer, Shahram Izadi, and
  Christian Theobalt.
\newblock Bundlefusion: Real-time globally consistent 3d reconstruction using
  on-the-fly surface re-integration.
\newblock {\em ACM Transactions on Graphics 2017 (TOG)}, 2017.

\bibitem{softposit}
Philip David, Daniel DeMenthon, Ramani Duraiswami, and Hanan Samet.
\newblock Softposit: Simultaneous pose and correspondence determination.
\newblock In Anders Heyden, Gunnar Sparr, Mads Nielsen, and Peter Johansen,
  editors, {\em Computer Vision --- ECCV 2002}, pages 698--714, Berlin,
  Heidelberg, 2002. Springer Berlin Heidelberg.

\bibitem{lsd_slam}
Jakob Engel, Thomas Sch{\"o}ps, and Daniel Cremers.
\newblock Lsd-slam: Large-scale direct monocular slam.
\newblock In David Fleet, Tomas Pajdla, Bernt Schiele, and Tinne Tuytelaars,
  editors, {\em Computer Vision -- ECCV 2014}, pages 834--849, Cham, 2014.
  Springer International Publishing.

\bibitem{omni_hard_why_2}
Hannes Fassold.
\newblock Adapting computer vision algorithms for omnidirectional video, 2019.

\bibitem{flownet}
Philipp Fischer, Alexey Dosovitskiy, Eddy Ilg, Philip Häusser, Caner
  Hazırbaş, Vladimir Golkov, Patrick van~der Smagt, Daniel Cremers, and
  Thomas Brox.
\newblock Flownet: Learning optical flow with convolutional networks, 2015.

\bibitem{ransac}
Martin~A. Fischler and Robert~C. Bolles.
\newblock Random sample consensus: A paradigm for model fitting with
  applications to image analysis and automated cartography.
\newblock {\em Commun. ACM}, 24(6):381--395, 1981.

\bibitem{omni_hard_why_4}
P. {Frossard} and R. {Khasanova}.
\newblock Graph-based classification of omnidirectional images.
\newblock In {\em 2017 IEEE International Conference on Computer Vision
  Workshops (ICCVW)}, pages 860--869, 2017.

\bibitem{pnp_1}
J.~A. {Hesch} and S.~I. {Roumeliotis}.
\newblock A direct least-squares (dls) method for pnp.
\newblock In {\em 2011 International Conference on Computer Vision}, pages
  383--390, 2011.

\bibitem{learned_loc_2}
Martin Humenberger, Yohann Cabon, Nicolas Gu{\'{e}}rin, Julien Morat,
  J{\'{e}}r{\^{o}}me Revaud, Philippe Rerole, No{\'{e}} Pion,
  C{\'{e}}sar~Roberto de Souza, Vincent Leroy, and Gabriela Csurka.
\newblock Robust image retrieval-based visual localization using kapture.
\newblock {\em CoRR}, abs/2007.13867, 2020.

\bibitem{flownet2}
Eddy Ilg, Nikolaus Mayer, Tonmoy Saikia, Margret Keuper, Alexey Dosovitskiy,
  and Thomas Brox.
\newblock Flownet 2.0: Evolution of optical flow estimation with deep networks,
  2016.

\bibitem{synthetic_view_with_matching}
A. {Irschara}, C. {Zach}, J. {Frahm}, and H. {Bischof}.
\newblock From structure-from-motion point clouds to fast location recognition.
\newblock In {\em 2009 IEEE Conference on Computer Vision and Pattern
  Recognition}, pages 2599--2606, 2009.

\bibitem{spatial_transformer}
Max Jaderberg, Karen Simonyan, Andrew Zisserman, and koray kavukcuoglu.
\newblock Spatial transformer networks.
\newblock In C. Cortes, N.~D. Lawrence, D.~D. Lee, M. Sugiyama, and R. Garnett,
  editors, {\em Advances in Neural Information Processing Systems 28}, pages
  2017--2025. Curran Associates, Inc., 2015.

\bibitem{omni_hard_why_1}
M. {Jayasuriya}, R. {Ranasinghe}, and G. {Dissanayake}.
\newblock Active perception for outdoor localisation with an omnidirectional
  camera.
\newblock In {\em 2020 IEEE/RSJ International Conference on Intelligent Robots
  and Systems (IROS)}, pages 4567--4574, 2020.

\bibitem{fukuoka}
H. {Jung}, Y. {Oto}, O.~M. {Mozos}, Y. {Iwashita}, and R. {Kurazume}.
\newblock Multi-modal panoramic 3d outdoor datasets for place categorization.
\newblock In {\em 2016 IEEE/RSJ International Conference on Intelligent Robots
  and Systems (IROS)}, pages 4545--4550, 2016.

\bibitem{posenet}
Alex Kendall, Matthew Grimes, and Roberto Cipolla.
\newblock Posenet: A convolutional network for real-time 6-dof camera
  relocalization.
\newblock 2015.

\bibitem{sgd}
J. Kiefer and J. Wolfowitz.
\newblock Stochastic estimation of the maximum of a regression function.
\newblock {\em Ann. Math. Statist.}, 23(3):462--466, 09 1952.

\bibitem{adam}
Diederik~P. Kingma and Jimmy Ba.
\newblock Adam: {A} method for stochastic optimization.
\newblock In Yoshua Bengio and Yann LeCun, editors, {\em 3rd International
  Conference on Learning Representations, {ICLR} 2015, San Diego, CA, USA, May
  7-9, 2015, Conference Track Proceedings}, 2015.

\bibitem{pnp_2}
Vincent Lepetit, Francesc Moreno-Noguer, and Pascal Fua.
\newblock Epnp: An accurate o(n) solution to the pnp problem.
\newblock {\em Int. J. Comput. Vision}, 81(2):155–166, Feb. 2009.

\bibitem{hierarchical_scene}
Xiaotian Li, Shuzhe Wang, Yi Zhao, Jakob Verbeek, and Juho Kannala.
\newblock Hierarchical scene coordinate classification and regression for
  visual localization.
\newblock In {\em CVPR}, 2020.

\bibitem{snavely}
Yunpeng Li, Noah Snavely, and Daniel~P. Huttenlocher.
\newblock Location recognition using prioritized feature matching.
\newblock In Kostas Daniilidis, Petros Maragos, and Nikos Paragios, editors,
  {\em Computer Vision -- ECCV 2010}, pages 791--804, Berlin, Heidelberg, 2010.
  Springer Berlin Heidelberg.

\bibitem{sift}
David~G. Lowe.
\newblock Distinctive image features from scale-invariant keypoints.
\newblock {\em International Journal of Computer Vision}, 60:91--110, 2004.

\bibitem{openmvg}
Pierre Moulon, Pascal Monasse, Romuald Perrot, and Renaud Marlet.
\newblock Openmvg: Open multiple view geometry.
\newblock In {\em International Workshop on Reproducible Research in Pattern
  Recognition}, pages 60--74. Springer, 2016.

\bibitem{dtam}
R.~A. {Newcombe}, S.~J. {Lovegrove}, and A.~J. {Davison}.
\newblock Dtam: Dense tracking and mapping in real-time.
\newblock In {\em 2011 International Conference on Computer Vision}, pages
  2320--2327, 2011.

\bibitem{pytorch}
Adam Paszke, Sam Gross, Francisco Massa, Adam Lerer, James Bradbury, Gregory
  Chanan, Trevor Killeen, Zeming Lin, Natalia Gimelshein, Luca Antiga, Alban
  Desmaison, Andreas Kopf, Edward Yang, Zachary DeVito, Martin Raison, Alykhan
  Tejani, Sasank Chilamkurthy, Benoit Steiner, Lu Fang, Junjie Bai, and Soumith
  Chintala.
\newblock Pytorch: An imperative style, high-performance deep learning library.
\newblock In H. Wallach, H. Larochelle, A. Beygelzimer, F. d\textquotesingle
  Alch\'{e}-Buc, E. Fox, and R. Garnett, editors, {\em Advances in Neural
  Information Processing Systems 32}, pages 8024--8035. Curran Associates,
  Inc., 2019.

\bibitem{learned_loc_1}
N. Pion, M. Humenberger, G. Csurka, Y. Cabon, and T. Sattler.
\newblock Benchmarking image retrieval for visual localization.
\newblock In {\em 2020 International Conference on 3D Vision (3DV)}, pages
  483--494, Los Alamitos, CA, USA, nov 2020. IEEE Computer Society.

\bibitem{learned_loc_3}
Paul{-}Edouard Sarlin, Cesar Cadena, Roland Siegwart, and Marcin Dymczyk.
\newblock From coarse to fine: Robust hierarchical localization at large scale.
\newblock {\em CoRR}, abs/1812.03506, 2018.

\bibitem{learned_loc_4}
Paul{-}Edouard Sarlin, Daniel DeTone, Tomasz Malisiewicz, and Andrew
  Rabinovich.
\newblock Superglue: Learning feature matching with graph neural networks.
\newblock {\em CoRR}, abs/1911.11763, 2019.

\bibitem{active_search_eccv}
Torsten Sattler, Bastian Leibe, and Leif Kobbelt.
\newblock Improving image-based localization by active correspondence search.
\newblock In Andrew Fitzgibbon, Svetlana Lazebnik, Pietro Perona, Yoichi Sato,
  and Cordelia Schmid, editors, {\em Computer Vision -- ECCV 2012}, pages
  752--765, Berlin, Heidelberg, 2012. Springer Berlin Heidelberg.

\bibitem{active_search}
Torsten Sattler, Bastian Leibe, and Leif Kobbelt.
\newblock Efficient \& effective prioritized matching for large-scale
  image-based localization.
\newblock {\em IEEE Trans. Pattern Anal. Mach. Intell.}, 39(9):1744--1756,
  2017.

\bibitem{change_vis_1}
T. {Sattler}, W. {Maddern}, C. {Toft}, A. {Torii}, L. {Hammarstrand}, E.
  {Stenborg}, D. {Safari}, M. {Okutomi}, M. {Pollefeys}, J. {Sivic}, F. {Kahl},
  and T. {Pajdla}.
\newblock Benchmarking 6dof outdoor visual localization in changing conditions.
\newblock In {\em 2018 IEEE/CVF Conference on Computer Vision and Pattern
  Recognition}, pages 8601--8610, 2018.

\bibitem{colmap_1}
Johannes~Lutz Sch\"{o}nberger and Jan-Michael Frahm.
\newblock Structure-from-motion revisited.
\newblock In {\em Conference on Computer Vision and Pattern Recognition
  (CVPR)}, 2016.

\bibitem{loc_feat_1}
Johannes~L. Schonberger, Hans Hardmeier, Torsten Sattler, and Marc Pollefeys.
\newblock Comparative evaluation of hand-crafted and learned local features.
\newblock In {\em Proceedings of the IEEE Conference on Computer Vision and
  Pattern Recognition (CVPR)}, July 2017.

\bibitem{colmap_2}
Johannes~Lutz Sch\"{o}nberger, Enliang Zheng, Marc Pollefeys, and Jan-Michael
  Frahm.
\newblock Pixelwise view selection for unstructured multi-view stereo.
\newblock In {\em European Conference on Computer Vision (ECCV)}, 2016.

\bibitem{score}
J. {Shotton}, B. {Glocker}, C. {Zach}, S. {Izadi}, A. {Criminisi}, and A.
  {Fitzgibbon}.
\newblock Scene coordinate regression forests for camera relocalization in
  rgb-d images.
\newblock In {\em 2013 IEEE Conference on Computer Vision and Pattern
  Recognition}, pages 2930--2937, 2013.

\bibitem{inloc}
Hajime Taira, Masatoshi Okutomi, Torsten Sattler, Mircea Cimpoi, Marc
  Pollefeys, Josef Sivic, Tomas Pajdla, and Akihiko Torii.
\newblock {InLoc: Indoor Visual Localization with Dense Matching and View
  Synthesis}.
\newblock In {\em {CVPR 2018 - IEEE Conference on Computer Vision and Pattern
  Recognition}}, Salt Lake City, United States, June 2018.

\bibitem{energy_landscape}
J. {Valentin}, A. {Dai}, M. {Niessner}, P. {Kohli}, P. {Torr}, S. {Izadi}, and
  C. {Keskin}.
\newblock Learning to navigate the energy landscape.
\newblock In {\em 2016 Fourth International Conference on 3D Vision (3DV)},
  pages 323--332, 2016.

\bibitem{lstm_vis_loc}
Florian Walch, Caner Hazirbas, Laura Leal-Taixe, Torsten Sattler, Sebastian
  Hilsenbeck, and Daniel Cremers.
\newblock Image-based localization using lstms for structured feature
  correlation.
\newblock In {\em Proceedings of the IEEE International Conference on Computer
  Vision (ICCV)}, Oct 2017.

\bibitem{change_vis_2}
Johanna Wald, Torsten Sattler, Stuart Golodetz, Tommasso Cavallari, and
  Frederico Tombari.
\newblock Beyond controlled environments: 3d camera re-localization in changing
  indoor scenes.
\newblock In {\em ECCV}, 2020.

\bibitem{sphere_cnn}
Chao Zhang, Ignas Budvytis, Stephan Liwicki, and Roberto Cipolla.
\newblock Rotation equivariant orientation estimation for omnidirectional
  localization.
\newblock In {\em ACCV}, 2020.

\bibitem{omni_hard_why_3}
Qiang Zhao, Chen Zhu, Feng Dai, Yike Ma, Guoqing Jin, and Yongdong Zhang.
\newblock Distortion-aware cnns for spherical images.
\newblock In {\em Proceedings of the Twenty-Seventh International Joint
  Conference on Artificial Intelligence, {IJCAI-18}}, pages 1198--1204.
  International Joint Conferences on Artificial Intelligence Organization, 7
  2018.

\bibitem{da4ad}
Yao Zhou, Guowei Wan, Shenhua Hou, L. Yu, Gang Wang, Xiaofei Rui, and Shiyu
  Song.
\newblock Da4ad: End-to-end deep attention-based visual localization for
  autonomous driving.
\newblock In {\em ECCV}, Aug 2020.

\bibitem{benefits_of_omni}
{Zichao Zhang}, H. {Rebecq}, C. {Forster}, and D. {Scaramuzza}.
\newblock Benefit of large field-of-view cameras for visual odometry.
\newblock In {\em 2016 IEEE International Conference on Robotics and Automation
  (ICRA)}, pages 801--808, 2016.

\end{thebibliography}

\end{document}